\documentclass[times,twocolumn,final]{elsarticle}

\usepackage{medima}
\usepackage{framed,multirow}

\usepackage{amssymb}
\usepackage{amsmath}
\usepackage{latexsym}
\usepackage{booktabs}

\usepackage{makecell}
\usepackage{threeparttable}

\usepackage{url}
\usepackage{xcolor}

\usepackage{hyperref}

\definecolor{newcolor}{rgb}{.8,.349,.1}

\journal{Medical Image Analysis}

\begin{document}

\verso{Shengyi Hua \textit{et~al.}}

\begin{frontmatter}

\title{PathoDuet: foundation models for pathological slide analysis of H\&E and IHC stains\tnoteref{tnote1}}%

\author[1]{Shengyi \snm{Hua}}
\author[2]{Fang \snm{Yan}}
\author[1]{Tianle \snm{Shen}}
\author[3]{Lei \snm{Ma}}
\author[1,2]{Xiaofan \snm{Zhang}\corref{cor1}}
\cortext[cor1]{Corresponding author.}
\ead{xiaofan.zhang@sjtu.edu.cn}

\address[1]{Qing Yuan Research Institute, Shanghai Jiao Tong University, Shanghai 200240, China}
\address[2]{Shanghai Artificial Intelligence Laboratory, Shanghai 200232, China}
\address[3]{National Biomedical Imaging Center, College of Future Technology, Peking University, Beijing 100871, China}

\begin{abstract}
    Large amounts of digitized histopathological data display a promising future for developing pathological foundation models via self-supervised learning methods. Foundation models pretrained with these methods serve as a good basis for downstream tasks. However, the gap between natural and histopathological images hinders the direct application of existing methods. In this work, we present PathoDuet, a series of pretrained models on histopathological images, and a new self-supervised learning framework in histopathology. The framework is featured by a newly-introduced pretext token and later task raisers to explicitly utilize certain relations between images, like multiple magnifications and multiple stains. Based on this, two pretext tasks, cross-scale positioning and cross-stain transferring, are designed to pretrain the model on Hematoxylin and Eosin (H\&E) images and transfer the model to immunohistochemistry (IHC) images, respectively. To validate the efficacy of our models, we evaluate the performance over a wide variety of downstream tasks, including patch-level colorectal cancer subtyping and whole slide image (WSI)-level classification in H\&E field, together with expression level prediction of IHC marker, tumor identification and slide-level qualitative analysis in IHC field. The experimental results show the superiority of our models over most tasks and the efficacy of proposed pretext tasks. The codes and models are available at \href{https://github.com/openmedlab/PathoDuet}{https://github.com/openmedlab/PathoDuet}.
\end{abstract}


\end{frontmatter}


\section{Introduction}
\label{sec:intro}

The histologic assessment stands as the gold standard for diagnosing specific cancers, predominantly depending on the expertise of pathologists. The assessment is mainly based on the analysis of Hematoxylin and Eosin (H\&E) stained slides, offering fundamental structural information. Pathologists may further augment their conclusions by utilizing functional stains such as Immunohistochemistry (IHC) to provide additional diagnostic insights. As technology continues to advance, digital scanners with high throughput have revolutionized the acquisition of pathological data. Despite large amounts of data, the integration of deep learning techniques into diagnostic processes has been progressing at a relatively measured pace. This can be attributed, in part, to the limited amount of labeled data for certain tasks. Unlike the annotation of natural images, the annotation process for pathological images demands expertise, rendering it resource-intensive and time-consuming. To address this challenge, foundation models emerge as a prospective solution in many medical fields \citep{ZHANG2024102996, Zhou2023, Li2023, wang2024foundationmodelendoscopyvideo, chen2023generalpurpose, vorontsov2023virchow, tiu2022expert, huang2023stunetscalabletransferablemedical, ma2024generalizablepathologyfoundationmodel}. These models typically exploit the potential of unlabeled data, facilitating efficient transferring to downstream tasks with reduced dependency on labeled data. 

Existing foundation models mainly rely on self-supervised learning (SSL) methodologies. 
The essence of SSL involves the generation of supervised signals directly from the data itself. This process is often called the pretext task \citep{jing2020self}. As a dominant branch of SSL methods, contrastive learning (CL) has attracted significant attention \citep{he2020momentum, chen2020simple, oquab2023dinov2}. In general, CL focuses on exploiting image similarity as a means to discern and categorize images concerning others. Another branch, featured by masked autoencoders \citep{he2022masked}, utilizes image generation to boost models' understanding. Compared with generative SSL methods, CL has better performance when transferred to discriminative tasks \citep{shekhar2023objectives}. As a result, CL is preferred in this work given the fact that quite a few pathological tasks are highly related to identification. 
However, the direct application of CL methods designed for natural images to histopathological images requires careful consideration. CL posits that the majority of images should possess semantic uniqueness. A routine is to contract different views of the same image and to separate different images in semantic space. Pathological whole slide images (WSI) are yet cropped into smaller patches to fit in the input size requirements of most models, restricting cropped patches to exhibit semantic distinctiveness from neighboring patches. Tough isolation of these patches may cause over-fragmentation of semantic space, thus affecting the performance of models. 
The conflict, consequently, requires a special design of contrastive pretext tasks concerning the essence of histopathological images. In seeking insights for the strategy of tailoring, inspiration can be drawn from the analogous process of pathological evaluation. 

A typical characteristic of pathologists' working methodology is their habitual practice of zooming in and out during the examination process. Initially, they employ a low magnification level to screen overall structures and tissues, identifying regions of interest that require closer inspection. Subsequently, at a higher magnification level, pathologists analyze individual cells or clusters of cells, refining their understanding and classification of the identified regions. 
To simulate the zooming in and out operations performed by pathologists, we define one of our pretext tasks as a ``cross-scale positioning" task, leveraging the large-scale public H\&E datasets to develop a foundation model for H\&E stain. 
In which, besides the commonly used two branches of different augmented image views in CL, we add a branch that learns the representation of a patch from its neighboring regions. 
This manipulation enables patches to be understood from a broader perspective, thereby alleviating the conflict between CL's requirement on semantic division and pathological patches' concentration. 

Additionally, pathologists often utilize additional functional slides for a more comprehensive diagnosis. Notably, IHC markers are frequently employed, offering valuable insights into subtyping cancers. However, the effective interpretation of IHC slides cannot be derived without H\&E stained slides. The H\&E slides serve as a fundamental reference, providing essential contextual information and structural details to complement the specific molecular information gleaned from IHC slides. 
Therefore, an ideal foundation model for IHC stain should be able to assess IHC images according to markers' expression levels and align with H\&E models in the semantic space in the aspect of the tissue structure. 
With the limited publicly available IHC data, we exploit the trained H\&E foundation model and introduce the ``cross-stain transferring" pretext task, to deepen comprehension of pathological images stained in a different way. 
Specifically, we align the IHC representation with the IHC-style transferred H\&E representation via a transferer drawing on the adaptive instance normalization, AdaIN \citep{huang2017arbitrary}. This alignment injects structural information that is readily accessible in H\&E images, as well as preserving diagnostic information rooted in IHC images.

\begin{figure*}
	\centering
	\includegraphics[width=0.85\textwidth]{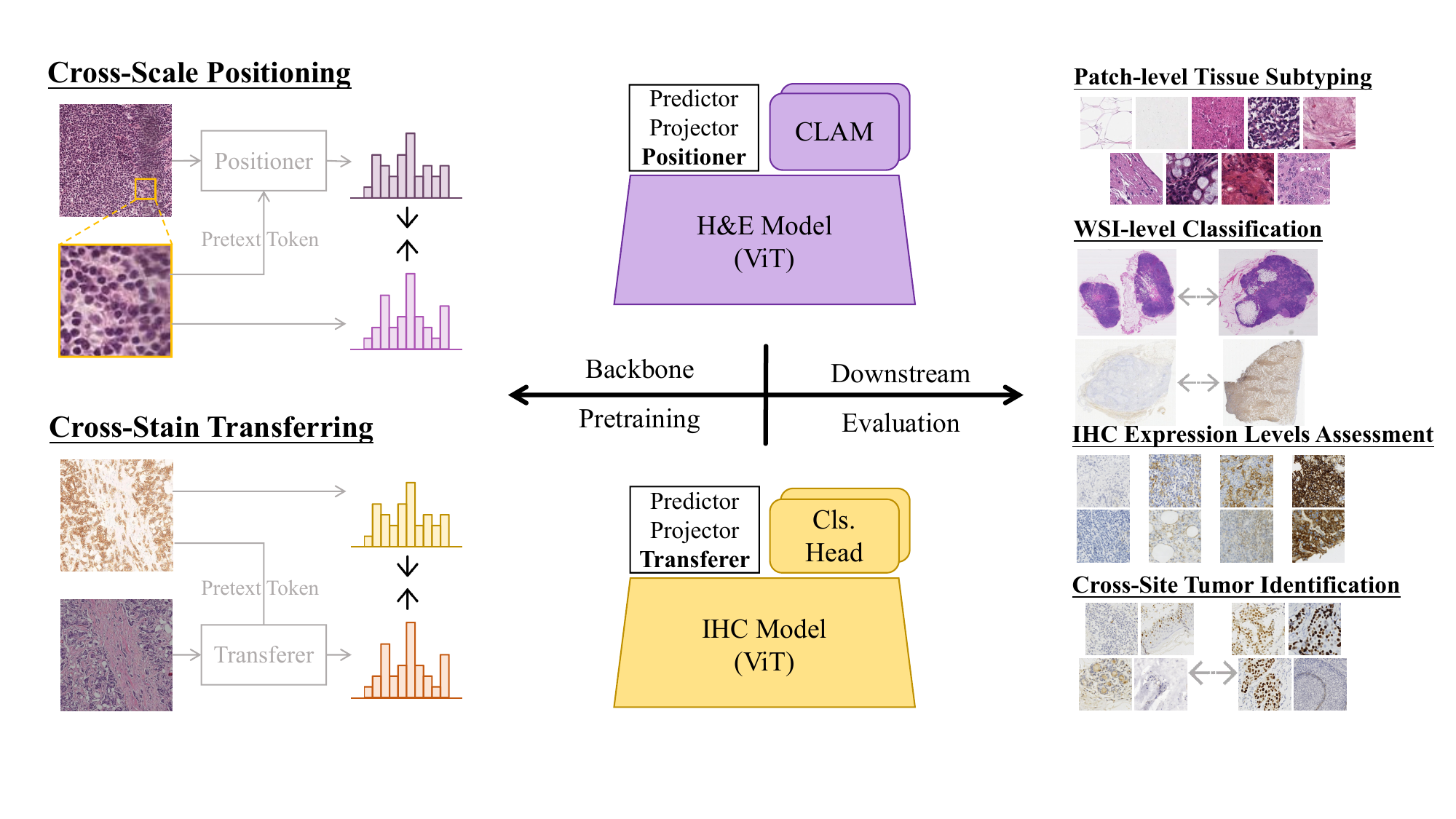}
	\caption{An overview of PathoDuet. Left: two pretext tasks, cross-scale positioning and cross-stain transferring, are designed to develop H\&E and IHC models. Right: a series of downstream tasks, covering both H\&E and IHC ones, are used to evaluate models' performance in application. }
	\label{fig:overall}
\end{figure*}

\begin{figure}
	\centering
	\includegraphics[width=0.85\columnwidth]{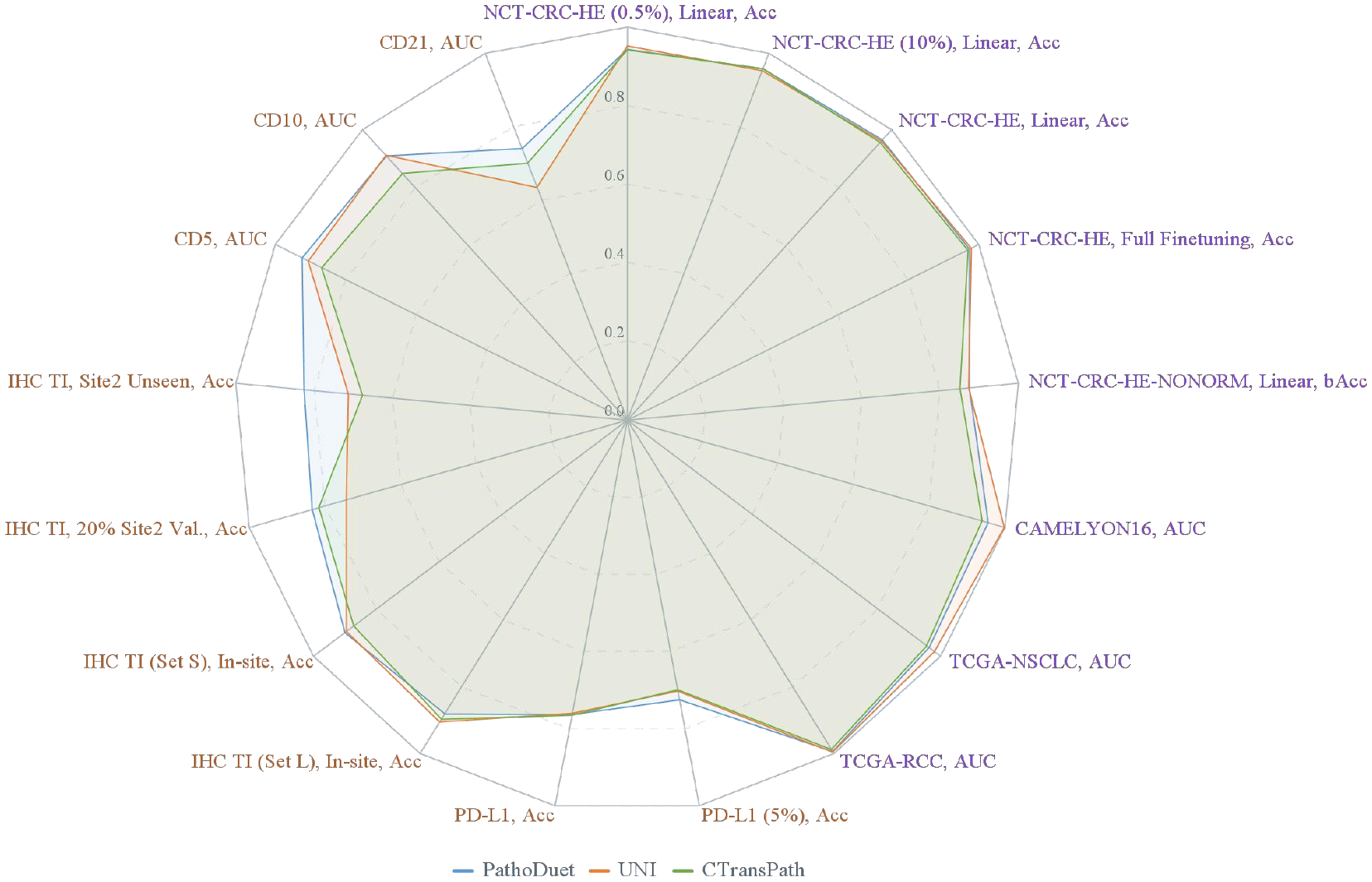}
	\caption{An overall performance visualization. Each task is named as \textit{training dataset, (special settings,) evaluating metric}. H\&E tasks are colored purple, and IHC ones are yellow.}
	\label{fig:radar}
\end{figure}
A pretext token mechanism is introduced to unify the two proposed tasks.
Both tasks require an auxiliary input in a different form, i.e., a much smaller patch or a staining hint. 
Contrary to designing a separate network to handle the additional input, an extra token with the auxiliary information is fed into the Vision Transformer (ViT) model \citep{dosovitskiy2020image} throughout the network training, and thereby combining the cross-scale or cross-stain information with original representation. The relation is subsequently exploited by a delicately designed module, termed a task raiser, to explicitly associate the two forms. This mechanism enriches the model's capacity to discover and leverage intrinsic correlation between tasks and staining modalities in a lightweight way.

The whole framework and ensuing models are collectively denoted as PathoDuet as shown in Fig. \ref{fig:overall}. This reflects the dual functionality of the framework, i.e., it offers two distinct strategies for developing foundation models: 1) by exploiting a shifted view of scales to discover broader semantic space, e.g., comprehending patches further from surrounding regions, 2) by progressing on the basis of other closely related and already exploited modalities, e.g., learning representation of IHC images from H\&E model. 
The efficacy of our proposed methods is validated across a wide range of downstream tasks, covering both H\&E and IHC images. \textcolor{black}{We use a spider chart in Fig. \ref{fig:radar} to visualize the overall performance across different tasks compared with powerful pathological models.} These tasks range from classifying cancer tissue types at the patch level, categorizing WSIs, identifying cancer cells within IHC images, assessing the expression levels of IHC markers, to WSI-level IHC qualitative analysis. Our contributions are summarized as: 

\begin{itemize}
    \item We introduce an auxiliary token into the plain ViT backbone, accompanied by task raisers for the designed pretext tasks, to build up a unified SSL framework. Within this framework, we propose PathoDuet, which is the first to provide both H\&E and IHC foundation models, to the best of our knowledge.
    \item We design the cross-scale positioning task to obtain a pretrained H\&E foundation model with a broader understanding, and the cross-stain transferring task to obtain an IHC interpreter from the existing H\&E foundation model with limited IHC data. 
    \item We validate the efficacy of our methodologies on several downstream tasks, consistently demonstrating superiority across the majority of these tasks. The codes and models are open-source to facilitate future use and reproductive experiments. 
\end{itemize}

\section{Related works}
\label{sec:related_work}

This section reviews the literature about SSL in computer vision and histopathology, respectively.

\subsection{Self-supervised learning}
Self-supervised learning can be seen as another learning strategy besides supervised and unsupervised learning. The difference from the supervised one is SSL requires no labeled data, so to this point, SSL is a special form of unsupervised learning. The difference from the purely unsupervised one is SSL requires supervision generated from the data, and the generating method is called pretext tasks. 

Pretext tasks can be various. Some tasks aim at predicting properties of images, like rotation angles \citep{gidaris2018unsupervised}, flipping \citep{srinidhi2022self}, etc. Some design small problems that help to learn features of images, like jigsaw puzzle solving \citep{noroozi2016unsupervised} and relative position prediction of sub-regions \citep{doersch2015unsupervised}. Some exploit the process of image generation, including image colorization \citep{zhang2016colorful}, super-resolution \citep{ledig2017photo}, inpainting \citep{pathak2016context}, and exploiting generative adversarial networks (GANs) \citep{zhu2017unpaired}. Besides, He et al. propose masked autoencoder (MAE) \citep{he2022masked} that leads recent research into masked image modeling.

Among all these pretext tasks, similarity-based CL task has demonstrated their superiority, because they focus on invariant features instead of covariant ones. CL-based methods pull together the latent representations of similar images while keeping those of dissimilar images away from each other. These methods often regard different augmented views from the same image as a similar (positive) pair, and those from different images as dissimilar (negative) pairs. Therefore, CL-based methods typically contain two network branches. MoCo v1 \citep{he2020momentum} is a basic one with a symmetric structure, a momentum-update mechanism, a memory bank, and an InfoNCE loss that considers positive samples and negative samples together. SimCLR \citep{chen2020simple} adds a simple multi-layer perceptron (MLP) called projector after each backbone encoder and updates two branches concurrently without a memory bank. BYOL \citep{grill2020bootstrap} further adds another MLP, predictor, after one of the branches (the online branch), while updating the other target branch with momentum, and changing the loss to cosine similarity loss which does not involve negative samples. SimSiam \citep{chen2021exploring} proposes a similar structure of BYOL but in a symmetric way with its gradient stop technique to avoid collapse. In MoCo v3 \citep{chen2021empirical}, it adopts the BYOL’s structure but continues using the InfoNCE loss. DINO \citep{caron2021emerging} and its later version DINO v2 \citep{oquab2023dinov2} integrate previous works and achieve another state-of-the-art performance, thereby attracting applications to other fields. 

\subsection{Self-supervised learning in digital pathology}
With the development of SSL frameworks in computer vision, the concept and some existing methods have been migrated to histopathology. Besides traditional pretext tasks, some works design specific pretext tasks in digital pathology. Magnification prediction \citep{sahasrabudhe2020self, koohbanani2021self} and stain prediction \citep{koohbanani2021self} are two simple tasks that exploit the characteristics of pathological images. Resolution sequence prediction \citep{srinidhi2022self} is a dedicated task with easy supervision and remarkable performance. Others leverage the stain. RestainNet \citep{zhao2022restainnet}, as its name suggests, simulates the process of de-staining and restaining via separated Hematoxylin and Eosin channels. \citet{ling2023self} use self-supervised methods to realize arbitrary stain transfer of pathological images from different domains. These methods though take into consideration the features of histopathological images, their performance depends on the association between the pretext tasks and the downstream tasks. 

Some works are based on CL frameworks. \citet{huang2021integration, li2021dual, ciga2022self, kawai2023large} directly migrate traditional CL methods to medical data, while some studies take into account the uniqueness of histopathological images and propose a modified SSL framework. CTransPath \citep{wang2022transformer} keeps the body of MoCo v3 but adds a pseudo positive selection mechanism to avoid false penalty of semantic similar patches and uses a hybrid CNN-Transformer as its backbone encoder. Meanwhile, Wang et al. also propose a clustering-guided contrastive learning method and the produced model RetCCL \citep{wang2023retccl}. CS-CO \citep{yang2022cs} separates histopathological images into H-channel and E-channel and uses a cross-stain prediction task in the first phase and a CL-based method in the second phase. \citet{abbet2022self} take advantage of domain information. Meanwhile, a branch of works has focused on the gigapixel WSIs \citep{vu2023handcrafted, wang2023dual, lazard2023giga, aryal2023context, schirris2022deepsmile, Xu2024, Li_2024_CVPR}. Recently, pathological models pretrained with ultra-large amounts of data have shown their superiority \citep{chen2023generalpurpose, dippel2024rudolfvfoundationmodelpathologists, vorontsov2023virchow}. \citet{chen2023generalpurpose} collect 100,000 slides to pretrain UNI with DINO v2 framework, and Virchow \citep{vorontsov2023virchow} with 1.5 million slides. Besides, some works look into multi-modal methods like visual-language learning and pretrain models on large-scale image-text datasets \citep{Huang2023PLIP, pisula2022language, lu2023towards, xu2024multimodalknowledgeenhancedwholeslidepathology}, and the focus is more on leveraging strong language models.

\section{Methods}
\label{sec:method}

In this section, we first describe the introduction of a pretext token and subsequent task raiser module to unify the proposed two pretext tasks. The details of the tasks are discussed in the following subsections, including the real-world inspiration and the imitation with the contrastive learning framework. 

\subsection{Pretext token empowered SSL framework}
The basic framework is based on a typical contrastive learning method, i.e., MoCo v3 \citep{chen2021empirical}. 
The difference starts from the input. To mimic pathologists, we need an extra input to bring the information from a different scale or stain. 
A typical practice to handle two input types is employing two networks respectively with a following interaction module. However, we aim at learning a single powerful encoder for the downstream tasks. 
Therefore, we propose the pretext token mechanism to uncover the relation within the same encoder, thus deepening understanding of pathological images like pathologists. 

The mechanism starts from adding a \emph{pretext token}, which is concatenated with the partitioned mini-patches' embeddings before being fed into the ViT. 
The token interacts with these embeddings and becomes an abstract association between the input pair in the attention blocks. After that, some simple and lightweight modules, termed as \emph{task raisers}, utilize the association to perform pretext tasks. As illustrated in Fig. \ref{fig:med}, the task raiser is instantiated as a \emph{positioner} to position a local patch from its surrounding region in the cross-scale positioning task, or a \emph{transferer} to generate IHC-style features from H\&E images in the cross-stain transferring task. 

In the following two subsections, we introduce in detail two pretext tasks we propose within this framework, and how pretext token and instantiated task raisers take effect in extracting information from two inputs. If not specified, we use the patch for the cropped image from a WSI and mini-patch for the output of patch embedding in ViT. Meanwhile, if the network takes only one kind of input, e.g., patches in patch network and IHC images in IHC network, the final input is formulated as $[\mathbf{x}; \epsilon]$, where $\mathbf{x}=[x_1; \dots; x_L] \in \mathbb{R}^{L\times C}$ represents the original input after patch embedding module in ViT. $\epsilon\in \mathbb{R}^{C}$ is a set of learnable parameters to hold the place for the pretext token so that the architecture of networks can be consistent and the usage of our models can be similar to that of a normal ViT. $L$ is the embedding length, and $C$ is the number of channels. When the network takes two inputs, the input changes to $[\mathbf{x}^o; x^p]$, where $\mathbf{x}^o=[x^o_1; \dots; x^o_L] \in \mathbb{R}^{L\times C}$ stands for the embeddings of the original input, e.g., a region or an H\&E image, and $x^p \in \mathbb{R}^{C}$ for those of the pretext token input, e.g., a patch or an IHC crop.  

\subsection{Cross-scale positioning}
\label{sec:pp}
\begin{figure*}
	\centering
	\includegraphics[width=0.85\textwidth]{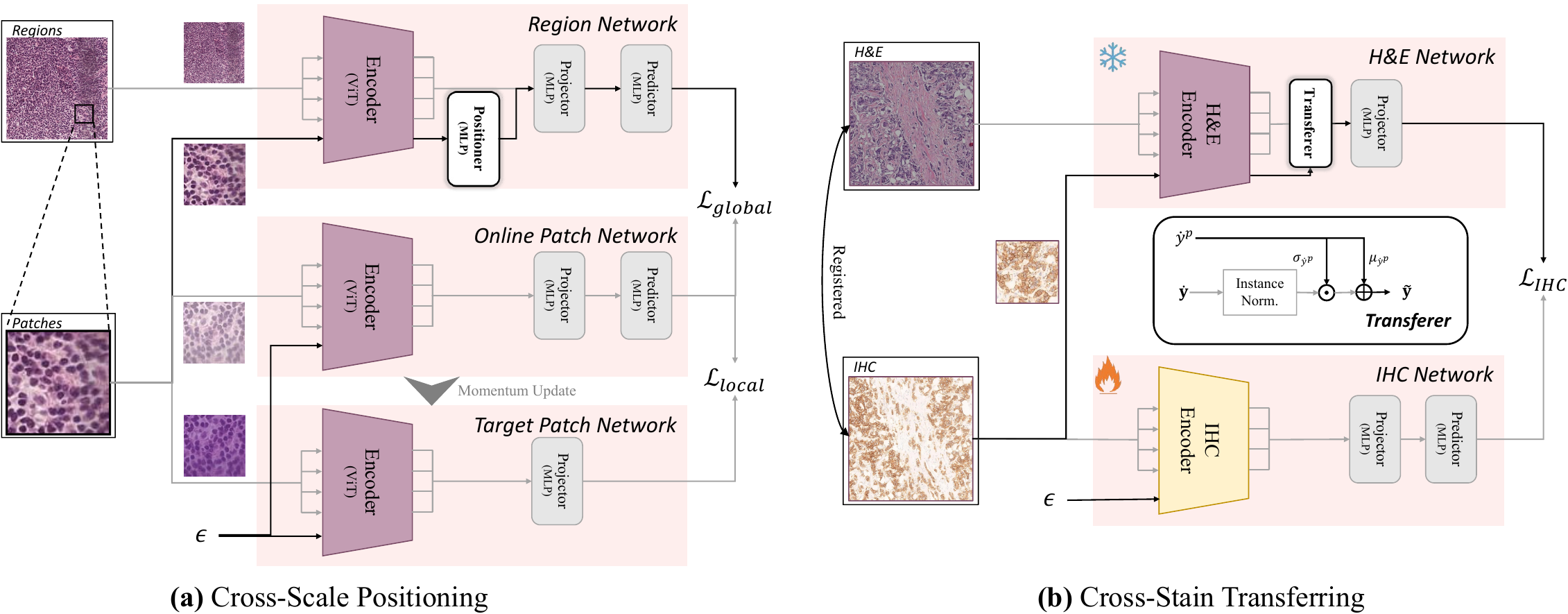}
	\caption{Detailed networks of two pretext tasks. The flow of pretext token is represented by the black arrows, $\epsilon$ is the placeholder, and the task raisers (positioner and transferer) are presented in the white blocks. (a) Three-branch cross-scale positioning network. (b) Two-branch cross-stain transferring network and the transferer module. }
	\label{fig:med}
\end{figure*}

If one observes how pathologists view slides, the most frequent action of them is likely to be zooming in and out. This phenomenon originates from the diagnostic process wherein pathologists first detect suspicious regions from a global view, then zoom in and define the region with its local surroundings, and subsequently zoom out to inspect the next region of interest. The underlying concept is that they obtain a global and coarse understanding of a WSI under low magnification, while a local but fine understanding of critical regions under high magnification. Inspired by this, we abstract the concept as a larger region containing a smaller patch, and design the cross-scale positioning task. \textcolor{black}{Due to the abundance of H\&E data, this task can be applied as a normal SSL method to pretrain a model, resulting in a three-branch architecture shown in Fig. \ref{fig:med} (a). Two of the branches exactly form the original CL framework to provide basic understanding, while the rest one performs the cross-scale positioning task to enhance understanding of different scales.} 

\textcolor{black}{In this task, the goal is to bridge the representations of a patch from a local view and a global one.} The local understanding of a patch is directly obtained from the output features of patch networks. To extract a global understanding, we jointly input patches (as pretext token) and their surrounding regions (as main input) into the region network. We instantiate the task raiser as an explicit positioner module to get positioning weights over mini-patches of regions, and use the weights to extract the desired global features. 

As shown in Fig. \ref{fig:med} (a), the whole framework has three branches, a region network, an online patch network, and a target patch network.  

\textbf{Patch in a local view.} The two patch networks constitute a traditional CL framework, i.e., MoCo v3, to provide a basic understanding of patches. The input is a differently augmented view of the patch for each branch. The backbone encoder is denoted as $F(\cdot)$. The outputs of the patch network's backbone are believed to represent the patch in a local view, as [$\mathbf{y}; y^p] = F\left([\mathbf{x}; \epsilon] \right)$. Then, a typical projector $G(\cdot)$ and predictor $H(\cdot)$ are used to generate the key $\mathbf{k}=G\left( \mathbf{y} \right)$ and query $\mathbf{q}=H\left( \mathbf{k} \right)$ as defined in MoCo v3. The contrastive loss is regarded as the loss in a local view $\mathcal{L}_{\text{local}}$, and is defined as, 

\begin{equation}
\mathcal{L}_{\text{local}}=-\log \frac{\exp{(\mathbf{q} \cdot \mathbf{k}_{+}/\tau)}}{\sum_{i=1}^B \exp{(\mathbf{q} \cdot \mathbf{k}_i/\tau)}},
\end{equation}

where $\mathbf{k}_{+}$ is the key of a positive sample (that from a different view of the same image), $B$ the batch size, and $\tau$ the temperature hyper-parameter. The parameters are $\theta$ for the online branch and $\xi$ for the target. $\theta$ is updated with gradient, and a momentum update mechanism  \citep{he2020momentum} is applied as $\xi \rightarrow m\xi+(1-m)\theta$, where $m$ is a momentum factor. 

\textbf{Patch in a global view.} The region network and the online patch network perform the cross-scale positioning task. In the patch networks, the pretext token is just a learnable parameter to occupy the space as mentioned before, while in the region network, the token is fed with referenced patches, thereby defining the input as $[\textbf{x}^o; x^p]$ with $\textbf{x}^o$ from regions and $x^p$ from patches. When the token is passed through the encoder, it interacts with mini-patches of the region, and the output $[\mathbf{\dot{y}}; \dot{y}^p]=F([\textbf{x}^o; x^p])$ is then regarded as the association between the patch and the region. A positioner $P(\cdot)$ of simple structure like MLP, serving as the task raiser, learns to generate positioning weights over mini-patches given the association as,
\begin{equation}
\begin{aligned}
&\mathbf{p}=[p_1,\dots,p_L]=\text{Softmax} \left(P\left(\dot{y}^p\right)\right), \\
\end{aligned}
\end{equation}

where $p_i \in \mathbb{R}$. 

The weighted feature $\hat{\mathbf{y}}=\sum_{j=1}^L p_{j} \dot{y}_{j}$ combines the inherent feature of a patch and correlating features from the corresponding region, thereby considered to be the patch representation in a global view. The key $\hat{\mathbf{k}}= G\left( \mathbf{\hat{y}} \right)$ and query $\mathbf{\hat{q}}=H\left( \mathbf{\hat{k}} \right)$ are simply acquired as before.

\textbf{Bridge local and global views.} The global feature and the local feature are then pulled together. Considering both the positioner and the patch network should be updated, we adopt SimSiam's two-way loss to propagate gradient in two networks, i.e., 
\begin{equation}
\mathcal{L}_{\text{global}}=D\left(\mathbf{\hat{q}}, \text{stopgrad}\left(\mathbf{k}\right)\right)+D\left(\mathbf{q}, \text{stopgrad}\left(\mathbf{\hat{k}}\right)\right),
\end{equation}
where $D(\mathbf{q}, \mathbf{k})=-\frac{\mathbf{q}}{\|\mathbf{q}\|_2} \cdot \frac{\mathbf{k}}{\|\mathbf{k}\|_2}$ is cosine similarity, and $\text{stopgrad}(\cdot)$ means a gradient stop manipulation \citep{chen2021exploring}. The local and global losses are then optimized equally. 

\subsection{Cross-stain transferring}
\label{sec:xst}
As mentioned earlier, pathologists cannot always give a detailed diagnosis if only H\&E slides are available. They sometimes require more evidence like IHC slides to define the disease. However, limited by the inadequacy of IHC slides, they should detect abnormal regions in H\&E slides, find the corresponding regions in IHC slides, and finally derive the expected information. The underlying concept is they obtain structural information from H\&E images and further diagnostic information from corresponding IHC regions. 
Inspired by this, we formulate the relation as paired H\&E and IHC images and design the cross-stain transferring task. \textcolor{black}{Considering the scarcity of desired datasets, and a priori structural knowledge in the H\&E pretrained model, this task is implemented in a two-branch way without a traditional CL branch and initiated from existing H\&E models. 
The whole process can be seen as we first acquire basic knowledge about pathological images using traditional CL method, then we enhance the global understanding using cross-scale positioning task, and now we use another cross-stain transferring task to explicitly translate H\&E understanding into IHC.} 

\textcolor{black}{In this task, the goal is to bridge the representations of a real IHC patch and a pseudo one transferred from the corresponding H\&E patch.} The real representations are drawn directly from the IHC network. As for the pseudo ones, we borrow the concept of adaptive instance normalization (AdaIN) \citep{huang2017arbitrary} that the style of images is rooted in some feature statistics like channel mean and variance, and utilize it to design a transferer as task raiser in the H\&E network. After we jointly input IHC references (as pretext token) and their paired H\&E patches (as main input) into the H\&E network, the transferer replaces the mean and standard error of H\&E features with those of IHC reference, thereby transferring H\&E features into IHC styles. The normalized features are viewed as pseudo features of the corresponding IHC patch with injected existing H\&E semantics, e.g., separation of structurally normal and abnormal patches. Meanwhile, a basic CL method with negative samples ensures learning the innateness of IHC images.  

As shown in Fig. \ref{fig:med} (b), the whole framework is a typical two-branch one, including an H\&E network and an IHC network. 

\textbf{Real IHC features.} The IHC network is initiated with previous H\&E models to provide a basic pathological understanding and gets updated with gradient during the transferring task. The network takes only IHC images and uses placeholders to occupy the pretext token. The outputs $[\mathbf{y}; y^p] = F\left([\mathbf{x}; \epsilon] \right)$ are believed to represent the basic understanding of IHC patches and used to compute the key $\mathbf{k}=G\left( \mathbf{y} \right)$ and query $\mathbf{q}=H\left( \mathbf{k} \right)$.  

\textbf{Pseudo IHC features.} The H\&E network is also initiated with the previous H\&E model but froze during this task to provide a stable understanding of H\&E images. The network takes both H\&E images as the main input and a cropped IHC reference as the pretext token. The crop is of moderate size of the original IHC image to provide adequate stain information while avoiding information leaks. The outputs of the network $[\mathbf{\dot{y}}; \dot{y}^p]=F([\textbf{x}^o; x^p])$ are instance normalized with the crop's feature via the transferer to provide a pseudo target for IHC features. 

\begin{equation}
\begin{aligned}
\tilde{\mathbf{y}} &= \text{AdaIN}\left(\mathbf{\dot{y}}, \dot{y}^p\right) \\
&= \frac{\mathbf{\dot{y}} - \mu_{\mathbf{\dot{y}}}}{\sigma_{\mathbf{\dot{y}}}} \cdot \sigma_{\dot{y}^p}+\mu_{\dot{y}^p},
\end{aligned}
\end{equation}
where $\mu_\mathbf{\dot{y}}$ and $\sigma_\mathbf{\dot{y}}$ represent the channel mean and standard error of $\mathbf{\dot{y}}$, and $\mu_{\dot{y}^p}$ and $\sigma_{\dot{y}^p}$ likewise. The key $\mathbf{\tilde{k}}=G\left( \mathbf{\tilde{y}} \right)$ is computed as before. 

\textbf{Bridge features.} The real IHC feature and the pseudo feature are then pulled together with typical InfoNCE loss $\mathcal{L}_{\text{IHC}}$, which is aware of negative samples. 

\begin{equation}
\mathcal{L}_{\text{IHC}}=-\log \frac{\exp{(\mathbf{q} \cdot \mathbf{\tilde{k}}_{+}/\tau)}}{\sum_{i=1}^B \exp{(\mathbf{q} \cdot \mathbf{\tilde{k}}_i/\tau)}},
\end{equation}

where $\mathbf{\tilde{k}}_{+}$ is the pseudo key of a positive sample, $B$ and $\tau$ the same as mentioned before.

\section{Experiments}
\label{sec:exp} 
In this section, we first introduce the datasets and experimental settings to obtain the H\&E and IHC models. Next, we describe in detail the downstream tasks of H\&E and IHC images sequentially. The H\&E tasks include patch-level colorectal cancer (CRC) tissue typing and WSI-level classification to evaluate both basic discriminating capability and global understanding of H\&E models. The IHC tasks include a typical assessment of the IHC marker's expression level, and a cross-site tumor identification to demonstrate basic IHC understanding, as well as generalization competence of models. 

These results are based on quantitative experiments with other models. The models first include models using ImageNet \citep{5206848} as baselines, i.e., fully-supervised ViT \citep{dosovitskiy2020image} denoted as \emph{ImageSup} and self-supervised ViT via MoCo v3 \citep{chen2021empirical} as \emph{ImageSSL}. Besides, some pathological models are also taken into consideration. \citet{ciga2022self} utilize SimCLR strategy to pretrain a ResNet-18 with large amounts of public pathological data (\emph{SimCLR-ciga}). Wang et al. provide both a pretrained ResNet-50, \emph{RetCCL} \citep{wang2023retccl}, and a pretrained hybrid architecture of CNN and Swin Transformer, \emph{CTransPath} \citep{wang2022transformer}, via customized methods and pathological datasets. The last is \emph{UNI} \citep{chen2023generalpurpose}, a ViT-Large pretrained with over 100 thousand slides using DINO v2 framework. These open-source models are compared throughout the experiments in this section. Pretrained model with ultra-large scale amounts of pathological data, like \emph{Virchow} \citep{vorontsov2023virchow}, is postponed to Section \ref{sec:dis-paige} for a simple discussion since it is not publicly available at this time. 

\subsection{Pretraining stage}
The pretraining stage consists of two stages, pretraining an H\&E model and transferring to an IHC model. In the first stage, we perform the cross-scale positioning task under the MoCo v3 framework with our H\&E dataset. In the next stage, we perform the cross-stain transferring task to the H\&E model with the cross-stain dataset.

\textbf{Datasets.} The \emph{H\&E dataset} originates from TCGA \footnote{\href{https://portal.gdc.cancer.gov/}{https://portal.gdc.cancer.gov/}}, a large-scale public dataset containing genome, epigenome, transcriptome, and image data. In this work, we collect about 30 thousand WSIs from it and select about 11 thousand diagnostic formalin-fixed paraffin-embedded (FFPE) WSIs for training. The patches are cropped under the highest magnification level with a size of $256 \times 256$ pixels, and the regions under the second highest level with a size of $1024 \times 1024$ pixels. Hence, the physical ratio of size between regions and patches is typically around 8. Finally, 1,623,258 regions and 13,166,437 patches are acquired. The \emph{cross-stain dataset} originates from HyReCo \citep{lotz2022comparison, pzj5-bs61-21} and BCI dataset \citep{Liu_2022_CVPR}. HyReCo dataset consists of nine datasets of consecutive sections, each containing four slides stained with H\&E, CD8, CD45, and Ki67, respectively. Additional PHH3-stained slides are re-stained from the bleached H\&E slides. In total, 2,771 pairs of H\&E and one stain of IHC are acquired. BCI dataset contains 4,873 pairs of H\&E and HER2 images, 3,896 pairs for train, and 977 for test. We only use the 3,896 training pairs. To obtain more training data, we crop these images under another resolution, and finally, 21,126 pairs are used in the second task.

\begin{table*}[t]
	\caption{Linear evaluation results on NCT-CRC-HE dataset with different amounts of training data. The best performance in each column is bold, and the second best is underlined.}
	\centering
	\begin{tabular}{lcccccccccc}
		\toprule
		\multirow{4}{*}{Methods} & \multicolumn{10}{c}{Percentage of training data}                   \\
		\cmidrule(r){2-11}
		   & \multicolumn{2}{c}{0.5$\%$} & \multicolumn{2}{c}{1$\%$} & \multicolumn{2}{c}{10$\%$} & \multicolumn{2}{c}{50$\%$} & \multicolumn{2}{c}{100$\%$} \\
    \cmidrule(r){2-3} \cmidrule(r){4-5} \cmidrule(r){6-7} \cmidrule(r){8-9} \cmidrule(r){10-11}
   & Acc&F1 &Acc &F1 &Acc &F1 &Acc &F1 &Acc &F1      \\
		\midrule
        ImageSSL  & 0.882 & 0.844 & 0.900 & 0.866 & 0.931 & 0.900 & 0.934 & 0.906 & 0.935 & 0.908    \\
        ImageSup  & 0.922 & 0.895 & 0.934 & 0.914 & 0.942 & 0.931 & 0.947 & 0.932 & 0.946 & 0.933       \\
		\midrule
        SimCLR-ciga  & 0.928 & 0.900 & 0.932 & 0.906 & 0.935 & 0.910 & 0.936 & 0.908 & 0.938 & 0.910       \\
        RetCCL  & 0.943 & 0.914 & 0.943 & 0.916 & 0.945 & 0.920 & 0.944 & 0.924 & 0.945 & 0.924       \\
	CTransPath & 0.942 & 0.907 & \underline{0.952}& 0.923 & \underline{0.958} & 0.935 &    
        0.956 & 0.932& 0.956& 0.932      \\
	{UNI} & {\textbf{0.952}} & {\textbf{0.934}} & {\textbf{0.953}}& {\textbf{0.935}} & {0.953} & {\underline{0.937}} &    
        {\underline{0.958}} & {\underline{0.938}}& {\underline{0.961}}& {\underline{0.945}}      \\
        \midrule
  \textbf{Ours (H\&E)} & \underline{0.943} & \underline{0.919} & 0.950 & \underline{0.929} & \textbf{0.959} & \textbf{0.942} & \textbf{0.964} & \textbf{0.949} & \textbf{0.964} & \textbf{0.950} \\
		\bottomrule
	\end{tabular}
	\label{tab:NCT-linear}
\end{table*}

\textbf{Settings.} In the training stage, we use a typical ViT-B/16 backbone and activate automatic mixed precision by PyTorch. For later consistency with the cross-scale positioning task, we use the ``avg" mode instead of ``token''. Also, considering the gap between ImageNet and our histopathological dataset, we ignore the normalization with ImageNet's mean and variance. Following MoCo v3, the projector is a three-layer perceptron, and the predictor a two-layer perceptron with a hidden dimension of $4096$. The data augmentation strategies include random cropping and scaling, Gaussian blur, color distortion, random flipping, following BYOL. The batch size is $2048$, $\tau$ is set as $0.2$, and the learning rate is initially $0.00015$ with a scaler of $\text{BatchSize}/256$ and updated using a cosine decay schedule with a long warm-up of $40$ epochs. AdamW is used as the optimizer with $0.1$ weight decay. 

We first perform the vanilla contrastive learning method with $100$ epochs. It takes about $300$ hours on 8 A100 GPUs. Then, we activate the positioning task. The positioner is also a two-layer perceptron with the same hidden dimension. We train an extra $20$ epochs with $10$ warm-up epochs, which takes another $100$ hours. Then, the pretrained H\&E backbone can be transferred or used directly for later downstream tasks with H\&E-stained images. 

After that, we perform the cross-stain transferring task based on the previous model. The batch size is $512$, the learning rate is $0.0002$, and the total number of epochs is $100$ with $20$ warm-up epochs. It takes about $6$ hours on 4 GeForce RTX 3090 GPUs. After this stage, the model can be applied to downstream tasks with certain IHC images.

\subsection{Downstream tasks with H\&E images}
To better evaluate the performance and generalizing ability of PathoDuet, we conduct a series of downstream experiments. The experiments start from H\&E-related ones. 

Cancer tissue identification is one of the most important tasks in computational pathology, especially with H\&E images. One formulation of this task is patch-level classification, or supervised image classification, which presents a relatively simple way to test models' basic understanding of H\&E patches. To both study the native ability of models and simulate the real scenarios, a linear probing strategy \citep{chen2020simple} and a full finetuning one are implemented to compare models' performance. 

Another one is WSI-level classification, or weakly-supervised classification, which closely resembles real-world scenarios and is more persuasive in demonstrating models' global understanding of pathological slides. This task is usually implemented in a multiple instance learning way, and we utilize the attention-based CLAM \citep{lu2021data} as the framework to perform the MIL process. 

We use these two tasks as an evaluation of our H\&E model. A more concrete description of each task is arranged in the following two subsections, while the detailed parameter settings are in the supplementary materials.

\subsubsection{Patch-level tissue subtyping}
\label{sec:exp-crc}
Identifying cancerous tissue is the main work of a pathologist, thereby playing a dominant role in computational pathology. The nature of this behavior is to recognize tissues and distinguish abnormal images. Hence, a feasible formulation is to crop the WSIs into patches and then classify them. On the other hand, patch-level classification or fully-supervised classification is one of the most basic evaluations of models. To this, this task is preferred as a good prologue of the comprehensive assessment. 

In this task, the pretrained model is used as a feature extractor of cropped patches, followed by a classifying layer. The experiments are conducted under two different strategies. First, the typical \emph{linear evaluation} strategy, i.e., only the newly-added linear layer gets updated while the rest part is frozen. Second, the \emph{full fine-tuning} strategy, which is more likely to be applied in practice. In this way, the pretrained model gets trained together with the linear classifier. Accuracy (Acc) and F1 score are used as the evaluating metrics. It is worth noting that to better demonstrate models’ performance on transfer learning, we use different proportions of training data in linear evaluation, and thus all the results are reproduced ones considering the randomness of data splits. 

\textbf{Datasets.} We use the \emph{NCT-CRC-HE} dataset \citep{kather_jakob_nikolas_2018_1214456} for the patch classification task. It is a collection of histopathology images specifically focused on CRC. It consists of 9 categories of tissue types, with one category representing normal tissues (NORM) and the remaining 8 categories representing colorectal cancer tissues, including adipose (ADI), background (BACK), debris (DEB), lymphocytes (LYM), mucus (MUC), smooth muscle (MUS), cancer-associated stroma (STR) and colorectal adenocarcinoma epithelium (TUM). A total number of 100,000 image patches with a size of $224 \times 224$ pixels at $0.5$ microns per pixel (mpp) are used as the training set, while 7,180 image patches are used as the testing set. All training patches are cropped from 86 H$\&$E WSIs and testing patches from 50 H$\&$E WSIs.

\begin{table}
    \caption{Results on NCT-CRC-HE dataset for 2 different strategies: linear evaluation, full fine-tuning. The best performance in each column is bold, and the second best is underlined.}
    \centering
    \begin{tabular}{lcccc}
    \toprule
    \multirow{3}{*}{Methods}   & \multicolumn{2}{c}{Linear evaluation} & \multicolumn{2}{c}{Full fine-tuning}  \\
    \cmidrule(r){2-3} \cmidrule(r){4-5}
    &Acc&F1 &Acc &F1  \\
    \midrule
    ImageSSL  & 0.935 & 0.908 & 0.958 & 0.945\\
    ImageSup & 0.946 & 0.933 & 0.960 & 0.947   \\
    \midrule
    SimCLR-ciga  & 0.938 & 0.910 & 0.937 & 0.924  \\
    RetCCL  & 0.945 & 0.924 & 0.950 & 0.936  \\
    
    CTransPath & 0.956 & 0.932  & 0.969 & 0.960\\
    {UNI} & {\underline{0.961}} & {\underline{0.945}}  & {\textbf{0.979}} & {\textbf{0.967}} \\
    \midrule
    \textbf{Ours (H\&E)} & \textbf{0.964} & \textbf{0.950} & \underline{0.973} & \underline{0.964} \\
    \bottomrule
    \end{tabular}
    \label{tab:full}
\end{table}

\begin{table*}[t]
    \caption{Results of weakly-supervised WSI classification on three public datasets. The best performance in each column is bold, and the second best is underlined.}
    \centering
    \begin{tabular}{lcccccc}
    \toprule
    \multirow{3}{*}{Methods}  & \multicolumn{2}{c}{CAM16} & \multicolumn{2}{c}{NSCLC} & \multicolumn{2}{c}{RCC}  \\
    \cmidrule(r){2-3} \cmidrule(r){4-5} \cmidrule(r){6-7}
   &Acc&AUC &Acc &AUC &Acc &AUC   \\
    \midrule
    TransMIL & 0.884& 0.931 & - & - & - & -  \\
    CLAM-SB  & 0.884 & 0.940 & 0.894&0.951 &0.929 &0.986  \\
    CLAM-SB+ImageSSL  & 0.861 & 0.915 & 0.883 &  0.949 & 0.924 & 0.989 \\
    CLAM-SB+ImageSup  & 0.853 & 0.890 & 0.870 & 0.940 & 0.943 & 0.990 \\
    \midrule
    CLAM-SB+SimCLR-ciga & 0.899 & 0.953 & 0.900 & 0.949 & 0.931 & 0.987 \\
    CLAM-SB+RetCCL & 0.868 & 0.919 & 0.851 & 0.927 & 0.932 & 0.987 \\
    CLAM-SB+CTransPath & 0.868 & 0.940 & 0.904 & 0.956  & 0.928 &0.988  \\
    {CLAM-SB+UNI} & {\textbf{0.984}} & {\textbf{0.999}} & {\textbf{0.934}} & {\textbf{0.980}}  & {\textbf{0.959}} & {\textbf{0.995}}  \\
    \midrule
    \textbf{CLAM-SB+Ours (H\&E)} & \underline{0.930} & \underline{0.956} & \underline{0.908} & \underline{0.963} & \underline{0.954} & \underline{0.993} \\
    \bottomrule
    \end{tabular}
    \label{tab:WSI}
\end{table*}

\textbf{Settings.} For linear evaluation on the NCT-CRC-HE dataset, following CTransPath, we use Adam as the optimizer with a batch size of $96$. The initial learning rate is set to $0.0003$. Data augmentations of random horizontal, vertical, and 90-degree flips and random cropping are used. For the full fine-tuning, the learning rate for the linear classifier is set to $0.0003$, while the learning rate for the rest part of the pretrained model is set to $0.00003$. The maximum number of epochs is set to $50$ for all models to converge.

\textbf{Results.} In Table \ref{tab:NCT-linear}, we evaluate our H\&E model using the linear probing method under different amounts of data. From the result, we can see that our model performs well across various amounts of training data over other pretrained models. Meanwhile, it can be observed that a generally consistent increasing trend exists with the growth of amounts of training data, but the difference is relatively small for most models. A further study is conducted in Section \ref{sec:dis-fdt} on the training data requirements of foundation models. Notably, the giant UNI shows a dominant performance when the training data is extremely limited, which demonstrates its general interpretability of pathological images. In Table \ref{tab:full}, we present the evaluation of models' performance under different training strategies using the whole NCT-CRC-HE dataset. The results demonstrate that the proposed model is a good interpreter of H\&E images under both a quick linear transferring manner and a thorough full fine-tuning protocol. The performance gain can be owed to the cross-scale positioning task, which enhances the model's understanding under a broader view. To verify the assumption, an ablating study is discussed in Section \ref{sec:dis-pp}. UNI also provides decent performance, which shows its great understanding in pathology and powerful ViT-Large architecture.

\subsubsection{WSI-level classification}
\label{sec:exp-he_wsi}
Another formulation of cancer identification is closely related to real-world scenarios. The processing unit is the WSIs instead of small patches. In practice, the WSI classification task is typically weakly-supervised, since only global annotations are given, and the WSI-level labels may correspond to only small regions within. For this reason, this task challenges models’ global understanding of pathological images. 

Recent works \citep{campanella2019clinical,shao2021transmil, lu2021data} have demonstrated the effectiveness of multiple instance learning (MIL) on weakly-supervised classification of WSIs. These MIL-based methods typically follow a two-step approach. First, WSIs are divided into smaller patches to generate patch-level features by utilizing a pretrained model. Second, to generate a slide-level prediction, patch-level features are aggregated using various feature fusion techniques, including recurrent neural network (RNN) or Transformer-based aggregators \citep{campanella2019clinical, shao2021transmil} and attention-based pooling \citep{lu2021data}. 

For pretrained models, we freeze the parameters and utilize the attention-based CLAM as the framework to perform the MIL process. We use Acc and area under the receiver operating characteristic curve (AUC) score to evaluate the WSI classification task. The AUC is calculated using the macro-averaged one when the number of classes is larger than 2. Besides the aforementioned methods, we further include the original CLAM \citep{lu2021data} and TransMIL \citep{shao2021transmil} as baselines. It is worth noting that except for TransMIL, other results are reproductive ones considering the update of datasets. 

\textbf{Datasets.} We evaluate this task on three WSI-level datasets: CAMELYON16 (CAM16), TCGA non-small cell lung cancer (NSCLC), and TCGA renal cell carcinoma (RCC). \emph{CAMELYON16} dataset was released as part of the Camelyon16 challenge \citep{bejnordi2017diagnostic}, which focused on two types of breast cancer classification: benign tissue and metastatic breast cancer. The dataset consists of a total of 399 Whole Slide Images, with 270 WSIs for training and 129 for testing. Although the dataset does provide detailed pixel-level annotations, in the context of weakly-supervised classification, we only utilize the global slide-level annotations, i.e., whether a WSI contains tumor areas or not. \emph{TCGA-NSCLC} dataset is derived from the TCGA database for two class subtyping: lung squamous cell carcinoma (TCGA-LUSC) and lung adenocarcinoma (TCGA-LUAD). It contains a total of 1053 diagnostic WSIs, including 512 LUSC slides from 478 cases and 541 LUAD slides from 478 cases. \emph{TCGA-RCC} dataset is another subset of TCGA which includes three subtypes of kidney tumor: kidney chromophobe renal cell carcinoma (TCGA-KICH), kidney renal clear cell carcinoma (TCGA-KIRC), and kidney renal papillary cell carcinoma (TCGA-KIRP). It contains a total of 940 diagnostic WSIs, including 121 KICH slides from 109 cases, 519 KIRC slides from 513 cases, and 300 KIRP slides from 276 cases.

\textbf{Settings.} For the weakly-supervised WSI classification task, following CLAM, we freeze our pretrained model and use the Adam as the optimizer with a batch size of 1 (WSI/bag) and a weight decay of $0.00001$. The learning rate is set to $0.0002$, and the epochs to $50$. For the CAMELYON16 dataset, we adopt the official data split in the CAMELYON16 challenge. For TCGA-NSCLC and TCGA-RCC datasets, we use 5-fold Monte Carlo cross-validation to obtain a more stable result. Each WSI is cropped into non-overlapping small patches after removing the background area with a filter of saturation less than 15 in the CAMELYON16 dataset, and 8 in the TCGA dataset. 

\textbf{Results.} In Table \ref{tab:WSI}, various methods are compared using three different public datasets. UNI achieves all best across these three datasets, which demonstrate its power in understanding pathological slides. Excluding UNI, our model shows great performance over these three datasets, e.g., +3.1\% accuracy gain in CAMELYON16, +0.3\% in TCGA-NSCLC, and +1.1\% in TCGA-RCC. This demonstrate that it is effective to use cross-scale positioning tasks to enhance the global understanding of pathological images. As for other models, pathological models surpass original CLAM and ImageNet models in most cases, but the superiority is not consistent. This might be credited to the strong generalizing ability and global understanding gained from ImageNet, especially when the visual encoder is frozen. 

\subsection{Downstream tasks with IHC images}
As for the other part of PathoDuet, to evaluate the IHC model's competence, we conduct three additional experiments that are closely related to real-world diagnosis, expression level assessment of certain markers, cancer cell identification and slide-level IHC qualitative analysis. 

Assessing the expression level of IHC markers is a primary work for pathologists to assess an IHC slide. We formulate this task as a classification between IHC patches of different expression levels to test models' basic capability of tackling IHC images. Cancer cell identification is also critical in the analysis of IHC images because only the expressed cancer cells are of diagnostic significance. It is yet hard for pathologists to find these cells in IHC slides alone since the structure of single cells is too vague to be recognized. Hence, this task serves as an advanced challenge to models' understanding of IHC images. Meanwhile, with data from two sites available, this task is intended as a cross-site one to better evaluate models' generalizing ability. To note, these tasks are implemented under linear protocol to focus on the inherent capabilities of models. Slide-level IHC qualitative analysis tests models' performance on WSI diagnosis. We collect three different markers, and exploit multiple instance learning method to ask models to give a positive or negative prediction of certain marker. A detailed description is also delivered in the following subsections.

\subsubsection{IHC expression levels assessment}
\label{sec:exp-pdl1}
Besides tumor cell identification, assessment of certain marker's expression levels plays a primary role in IHC diagnosis. We formulate it as a patch-level multi-class classification task, because trivial regression may lose focus on certain scores that have diagnostic meaning. We manually selected several thresholds closely related to pathologists' examination. A linear probing method is applied with metrics including accuracy (Acc), balanced accuracy (bAcc), and weighted F1 (wF1) score because of class imbalance.

\textbf{Datasets.} Considering the scarcity of public datasets that meet the requirements, we use an in-house dataset. We collect two groups of IHC patches with PD-L1 marker from the same medical center. After exhaustive annotation of expression scores, we select 0.05, 0.2, and 0.5 as thresholds, thereby creating a 4-class classification task of rarely-, lightly-, moderately- and severely-expressed patches. In the first group, there are 765/1,059/645/481 patches in the order of expression level, and 693/1,138/491/478 patches in the second. We use Group 1 as the training set and Group 2 as the validation set. Besides the original setting, a more difficult setting is also implemented that only 5\% of training data is available, i.e., 38/46/38/25 patches respectively. 

\begin{table}
    \caption{Results of PD-L1 expression level assessment. The best performance in each column is bold, and the second best is underlined.}
    \centering
    \begin{tabular}{m{2cm}m{0.65cm}<{\centering}m{0.65cm}<{\centering}m{0.65cm}<{\centering}m{0.65cm}<{\centering}m{0.65cm}<{\centering}m{0.65cm}<{\centering}}
    \toprule
    \multirow{3}{*}{Methods}  & \multicolumn{3}{c}{100\% Training Set} & \multicolumn{3}{c}{5\% Training Set}  \\
    \cmidrule(r){2-4} \cmidrule(r){5-7}
    & Acc&bAcc&wF1 &Acc&bAcc&wF1  \\
    \midrule
    ImageSSL  & 0.754 & 0.721 & 0.753 & 0.686 & 0.698 & 0.695 \\
    ImageSup  & 0.726 & 0.688 & 0.715 & 0.648 & 0.653 & 0.651  \\
    \midrule
    SimCLR-ciga  & 0.754 & 0.754 & 0.744 & \underline{0.705} & 0.704 & \underline{0.710} \\
    RetCCL   & 0.751 & 0.754 & 0.754 & 0.677 & 0.693 & 0.686  \\
    
    CTransPath   & \textbf{0.765} & \textbf{0.762} & \textbf{0.768} & 0.700 & \underline{0.709} & 0.703 \\
    {UNI}   & {0.760} & {0.747} & {0.755} & {0.703} & {\underline{0.709}} & {0.701} \\
    \midrule
    \textbf{Ours (IHC)}   & \underline{0.763} & \underline{0.755} & \underline{0.765} & \textbf{0.726} & \textbf{0.721} & \textbf{0.732}  \\
    \bottomrule
    \end{tabular}
    \label{tab:pdl1}
\end{table}

\textbf{Settings.} For IHC expression levels assessment, we also keep the linear classification setting in Section \ref{sec:exp-crc}. The batch size is $128$ for full training data and $64$ for 5\% training data. The learning rate is $0.02$.

\textbf{Results.} In Table \ref{tab:pdl1}, the performance of different models is reported in different amounts of training data. 
Reviewing the overall results, we can suppose that pathological foundation models of H\&E images can have a certain insight into IHC images because we can see superior performance over ImageNet models. When focusing on individual models, our IHC model demonstrates attracting performance on most metrics, especially in the limited training data case. Notably, CTransPath also presents excellent performance, especially when training data is sufficient, and SimCLR-ciga provides second best performance when training data is limited. This might be credited to their use of some IHC images as pretraining data. Besides, although UNI does not see IHC images during pretraining, it still presents surprisingly powerful generalizing ability to IHC images. When the difficulty of the task increases, the advantage of explicit transferring to the IHC model is more obvious. The phenomenon states that although H\&E models can provide satisfactory results with adequate training data, explicit transferring is necessary for limited annotated data.

\begin{table*}
    \caption{Results of patch-level tumor identification in IHC images. The best performance in each column is bold, and the second best is underlined.}
    \centering
    \begin{tabular}{m{2cm}m{1.2cm}<{\centering}m{1.2cm}<{\centering}m{1.2cm}<{\centering}m{1.2cm}<{\centering}m{1.2cm}<{\centering}m{1.2cm}<{\centering}m{1.2cm}<{\centering}m{1.2cm}<{\centering}}
    \toprule
    \multirow{4}{*}{Methods}  & \multicolumn{4}{c}{Site 1 $\rightarrow$ Site 1}  &   \multicolumn{4}{c}{Site 1$\rightarrow$ Site 2}  \\
    \cmidrule(r){2-5} \cmidrule(r){6-9}
    & \multicolumn{2}{c}{Set L $\to$ Set S}  & \multicolumn{2}{c}{Set S $\to$ Set L} & \multicolumn{2}{c}{Site 2 Seen} & \multicolumn{2}{c}{Site 2 Unseen} \\
    \cmidrule(r){2-3} \cmidrule(r){4-5} \cmidrule(r){6-7} \cmidrule(r){8-9}
    & Acc&F1 &Acc&F1 &Acc&F1 &Acc&F1 \\
    \midrule
    ImageSSL & 0.817 & 0.811 & 0.864 & 0.862 & 0.547 & 0.545 & 0.504  & 0.503 \\
    ImageSup  & 0.881 & 0.879 & 0.875 & 0.875 & 0.677 & 0.667  & 0.537  & 0.535    \\
    \midrule
    SimCLR-ciga  & 0.849 & 0.846 & 0.766 & 0.754 & 0.762 & 0.744 & \underline{0.728} & \underline{0.717}  \\
    RetCCL   & 0.802 & 0.793  & 0.818 &  0.811 & 0.727 & 0.717 & 0.629 & 0.625 \\
    
    CTransPath   & \underline{0.897} & \underline{0.896}  & 0.872 & 0.870 & \underline{0.816} & \underline{0.794} & 0.677  & 0.657  \\
    {UNI}   & {\textbf{0.905}} & {\textbf{0.904}}  & {\underline{0.895}} & {\underline{0.894}} & {0.743} & {0.700} & {0.712}  & {0.692}  \\
    \midrule
    \textbf{Ours (IHC)}   & 0.881 & 0.881  & \textbf{0.900} &  \textbf{0.900} & \textbf{0.833} & \textbf{0.797} & \textbf{0.826} & \textbf{0.769} \\
    \bottomrule
    \end{tabular}
    \label{tab:ihc_cancer}
\end{table*}

\begin{table*}
    \caption{Results of slide-level prediction of CD5. The best performance in each column is bold, and the second best is underlined.}
    \centering
    \begin{tabular}{m{2cm}m{1.5cm}<{\centering}m{1.5cm}<{\centering}m{1.5cm}<{\centering}m{1.5cm}<{\centering}m{1.5cm}<{\centering}m{1.5cm}<{\centering}}
    \toprule
    Methods  & AUC  &   Acc & F1 & Recall & Precision & Specificity \\
    \midrule
    ImageSSL & \textbf{0.931} & \textbf{0.888} & \textbf{0.908} & \textbf{0.909} & \textbf{0.908} & \textbf{0.860} \\
    ImageSup  & \underline{0.924} & \underline{0.873} & \underline{0.895} & \underline{0.895} & \underline{0.899} & 0.844 \\
    \midrule
    SimCLR-ciga  & 0.862 & 0.798 & 0.837 & 0.850 & 0.826 & 0.722 \\
    RetCCL   & 0.881 & 0.825 & 0.852 & 0.840 & 0.866 & 0.803  \\
    
    CTransPath   & 0.868 & 0.814 & 0.844 & 0.834 & 0.856 & 0.784 \\
    UNI   & 0.907 & 0.863 & 0.887 & 0.888 & 0.887 & 0.828 \\
    \midrule
    \textbf{Ours (IHC)}   & \underline{0.924} & 0.840 & 0.862 & 0.835 & 0.895 & \underline{0.850} \\
    \bottomrule
    \end{tabular}
    \label{tab:ihc_markers1}
\end{table*}

\subsubsection{Cross-site tumor identification} 
\label{sec:exp-ihc}
As mentioned earlier, identifying tumor cells in IHC images is of high significance, but difficult for pathologists without auxiliary H\&E images. Therefore, this task can be a further examination of the models' abilities. Meanwhile, with the help of data from two different sites, we can further investigate models' generalizing competence under out-of-distribution settings. We formulate it as a patch-level classification task and use the linear protocol as well. The metrics are the Acc and F1 score. 

\textbf{Datasets.} The dataset is also private, consisting of IHC images from two medical sites. The annotation is simply positive (images containing tumor cells) and negative (images without any tumor cells). The first site (\emph{Site 1}) contains two sets of data stained with ER, KI67, and PR acquired at different times, resulting in slightly different appearances. The first set has 1,365 patches (700 positive and 665 negative), and the other has 126 patches (64 positive and 62 negative), denoted as \emph{Set L} and \emph{Set S} respectively. In the second site (\emph{Site 2}), there are 3,688 patches with much more domain shift, including 2,684 positive and 1,004 negative stained with ER, KI67, and EGFR. 

To better evaluate the models' performance, we design four scenarios. \emph{Set L $\rightarrow$ Set S} and \emph{Set S $\rightarrow$ Set L} are viewed as in-site settings. The difference is we use a larger dataset (Set L) as the training set in the former one and a smaller dataset (Set S) in the latter one. Data in Site 2 can be viewed as being sampled from a new domain other than Site 1. To this, it can be used for out-of-distribution (OOD) evaluation. We also design two OOD settings. \emph{Site 2 Seen} keeps 20\% of data in Site 2, and uses it for model selection. The model is trained with Set L in Site 1, validated with 20\% data in Site 2, and tested with the rest 80\% data. \emph{Site 2 unseen} is a stricter setting, which is also common in real-world scenarios. The model is trained with Set L in site 1, validated with Set S, and tested with data from absolutely unseen Site 2.

\textbf{Settings.} For IHC tumor identification, a linear probing method is implemented as well. In \emph{Set L $\to$ Set S}, \emph{Site 2 Seen} and \emph{Site 2 Unseen}, the batch size is $128$ and the learning rate is $0.005$. In \emph{Set S $\to$ Set L}, the batch size is reduced to $64$. The maximum number of epochs is also set to $50$ for convergence of all models.

\textbf{Results.} In Table \ref{tab:ihc_cancer}, the performance of different models is reported under four evaluation scenarios. When applying in-site settings, ImageNet supervised ViT, CTransPath, UNI, and our IHC model all present remarkable performance. When tested on another site, ImageNet models yet display a weakening performance. Meanwhile, our model offers the best performance in a setting of partially available data from a new domain and maintains the performance in a pure OOD setting. Other pathological models perform well with some a priori information from another domain, but may not keep it when the test set is totally unseen.

\begin{table*}
    \caption{Results of slide-level prediction of CD10. The best performance in each column is bold, and the second best is underlined.}
    \centering
    \begin{tabular}{m{2cm}m{1.5cm}<{\centering}m{1.5cm}<{\centering}m{1.5cm}<{\centering}m{1.5cm}<{\centering}m{1.5cm}<{\centering}m{1.5cm}<{\centering}}
    \toprule
    Methods  & AUC  &   Acc & F1 & Recall & Precision & Specificity \\
    \midrule
    ImageSSL & 0.898 & \textbf{0.842} & \textbf{0.813} & \underline{0.802} & \underline{0.827} & 0.873 \\
    ImageSup  & 0.870 & 0.800 & 0.758 & 0.737 & 0.784 & 0.848 \\
    \midrule
    SimCLR-ciga  & 0.902 & \textbf{0.842} & \underline{0.810} & \textbf{0.814} & 0.824 & 0.874 \\
    RetCCL   & 0.855 & 0.782 & 0.742 & 0.749 & 0.752 & 0.819  \\
    
    CTransPath   & 0.849 & 0.773 & 0.729 & 0.720 & 0.754 & 0.825 \\
    UNI   & \textbf{0.912} & 0.828 & 0.786 & 0.757 & 0.820 & \underline{0.880} \\
    \midrule
    \textbf{Ours (IHC)}   & \underline{0.909} & \underline{0.835} & 0.795 & 0.766 & \textbf{0.838} & \textbf{0.894} \\
    \bottomrule
    \end{tabular}
    \label{tab:ihc_markers2}
\end{table*}

\begin{table*}
    \caption{Results of slide-level prediction of CD21. The best performance in each column is bold, and the second best is underlined.}
    \centering
    \begin{tabular}{m{2cm}m{1.5cm}<{\centering}m{1.5cm}<{\centering}m{1.5cm}<{\centering}m{1.5cm}<{\centering}m{1.5cm}<{\centering}m{1.5cm}<{\centering}}
    \toprule
    Methods  & AUC  &   Acc & F1 & Recall & Precision & Specificity \\
    \midrule
    ImageSSL & 0.677 & 0.614 & \underline{0.667} & \underline{0.690} & 0.662 & 0.535 \\
    ImageSup  & 0.638 & 0.571 & 0.614 & 0.628 & 0.631 & 0.520 \\
    \midrule
    SimCLR-ciga  & 0.659 & 0.582 & 0.638 & 0.660 & 0.627 & 0.492 \\
    RetCCL   & 0.657 & 0.604 & 0.659 & 0.688 & 0.644 & 0.507  \\
    
    CTransPath   & \underline{0.701} & \textbf{0.637} & \underline{0.667} & 0.651 & \textbf{0.703} & \textbf{0.631} \\
    UNI   & 0.635 & 0.607 & 0.644 & 0.641 & 0.658 & \underline{0.570} \\
    \midrule
    \textbf{Ours (IHC)}   & \textbf{0.740} & \underline{0.626} & \textbf{0.669} & \textbf{0.695} & \underline{0.667} & 0.548 \\
    \bottomrule
    \end{tabular}
    \label{tab:ihc_markers3}
\end{table*}

\subsubsection{Qualitative analysis of IHC slides}
\label{sec:exp-ihc2}

Besides patch-level IHC tasks, diagnosis directly from IHC slides is also of great importance. Therefore, we collect some IHC slides of different markers, and invite some experts to give a positive or negative label to each slide. This task further examines models' capability of assessing IHC slides given certain marker. We still apply CLAM as the training method. The evaluating metrics include AUC, Acc, F1 score, Recall, Precision and Specificity. 

\textbf{Datasets.} The dataset includes 3 IHC markers, CD5, CD10 and CD21. For each marker, we collect over 250 slides and annotate each slide with a positive or negative label. To be detailed, we collect 124/189 positive/negative CD5 slides, 139/111 CD10 slides, and 115/150 CD21 slides, respectively. We use each marker solely and implement 5-fold cross validation. 

\textbf{Settings.} For slide-level prediction of IHC markers, we follow the settings used in TCGA experiments as introduced in Section \ref{sec:exp-he_wsi}.

\textbf{Results.} In Table \ref{tab:ihc_markers1}, \ref{tab:ihc_markers2} and \ref{tab:ihc_markers3}, we present the results of slide-level prediction of CD5, CD10 and CD21. If we take three tables as a whole, we can see that ImageSSL, SimCLR-ciga, CTransPath, UNI and our models provide relatively good and stable performance. For ImageSSL, an interesting fact is that it surpasses ImageSup in all statistics. A reasonable guess is self-supervised learning helps model generalizing to other domains, especially those rarely-seen domains. For SimCLR-ciga and CTransPath, as aforementioned, they may benefit from some IHC data used for pretraining. For UNI, it still presents great performance with its surprising generalizing ability. For our model, it provides outstanding performance, achieving at least second-best AUC across all three datasets, while no other models can achieve. This should be credited to cross-stain transferring, which exploits existing H\&E model and a little amount of paired IHC and H\&E data. Reviewing the overall results, we can derive similar conclusions. Foundation models of H\&E images can be a good choice to understand IHC images, compared with models trained with natural images. An IHC-specific model, however, can provide more insights especially when the difficulty of the task is relatively high, i.e., large domain shift and limited annotated data.

\section{Discussion}
\label{sec:dis}
In this section, we first conduct several ablation experiments to validate the efficacy of two proposed pretext tasks. Then, we study the data requirement of applying foundation models to downstream tasks, to demonstrate the necessity of developing pathological foundation models. Finally, we compare our H\&E model to some giant models pretrained with ultra-large scale pathological datasets and observe that both increasing the amount of data and designing a tailored framework help to develop powerful pathological models. 

\begin{table}
    \caption{Ablation study: performance on WSI classification.}
    \centering
    \begin{tabular}{lm{0.7cm}<{\centering}m{0.7cm}<{\centering}m{0.7cm}<{\centering}m{0.7cm}<{\centering}m{0.7cm}<{\centering}m{0.7cm}<{\centering}}
    \toprule
    \multirow{3}{*}{Model} & \multicolumn{2}{c}{CAM16} & \multicolumn{2}{c}{NSCLC} & \multicolumn{2}{c}{RCC}  \\
    \cmidrule(r){2-3} \cmidrule(r){4-5} \cmidrule(r){6-7}
    &Acc&AUC &Acc  &AUC &Acc  &AUC  \\
    \midrule
    MoCo v3   & 0.915 & 0.959 & 0.890 & 0.937 & 0.948 & 0.992  \\
    +XSP & 0.930  & 0.956 & 0.908  & 0.963 & 0.954  & 0.993 \\
    \bottomrule
    \end{tabular}
    \label{tab:ab1}
\end{table}

\begin{table}
    \caption{Ablation study: performance on H\&E patch classification.}
    \centering
    \begin{tabular}{lcccc}
    \toprule
    \multirow{3}{*}{Model} & \multicolumn{2}{c}{Linear evaluation} & \multicolumn{2}{c}{Full fine-tuning}  \\
    \cmidrule(r){2-3} \cmidrule(r){4-5}
    &Acc&F1 &Acc  &F1  \\
    \midrule
    MoCo v3 &  0.956 & 0.944 & 0.973 & 0.960  \\
    +XSP & 0.964  & 0.950 & 0.973  & 0.964 \\
    \bottomrule
    \end{tabular}
    \label{tab:ab2}
\end{table}

\begin{table}
    \caption{Ablation study: performance on PD-L1 expression level assessment.}
    \centering
    \begin{tabular}{lm{0.7cm}<{\centering}m{0.7cm}<{\centering}m{0.7cm}<{\centering}m{0.7cm}<{\centering}m{0.7cm}<{\centering}m{0.7cm}<{\centering}}
    \toprule
    \multirow{3}{*}{} & \multicolumn{3}{c}{100\% Training Set} & \multicolumn{3}{c}{5\% Training Set}  \\
    \cmidrule(r){2-4} \cmidrule(r){5-7}
    &Acc&bAcc&wF1 &Acc &bAcc &wF1  \\
    \midrule
    H\&E  & 0.762 & 0.763 & 0.764 & 0.714 & 0.699 & 0.717  \\
    IHC & 0.763 & 0.755 & 0.765 & 0.726 & 0.721 & 0.732 \\
    \bottomrule
    \end{tabular}
    \label{tab:ab3}
\end{table}

\begin{table*}
    \caption{Comparative study on NCT-CRC-HE and NCT-CRC-HE-NONORM dataset. SwinT* means a hybrid model of CNN and Swin Transformer. The best performance in each column is bold, and the second best is underlined.}
    \centering
    \begin{tabular}{lllcccccc}
    \toprule
    \multirow{3}{*}{} & \multirow{3}{*}{Architecture} & \multirow{3}{*}{\#WSIs} & \multicolumn{3}{c}{CRC} & \multicolumn{3}{c}{CRC (no norm)}  \\
    \cmidrule(r){4-6} \cmidrule(r){7-9}
    &&&Acc&bAcc&wF1 &Acc&bAcc&wF1  \\
    \midrule
    Ours (H\&E) & \multirow{2}{*}{ViT-Base} & \multirow{2}{*}{$\sim$ 11K} & \underline{0.964} & \underline{0.952} & \underline{0.964} & 0.888 & 0.875 & 0.894 \\
    Ours (H\&E)* &  &  & - & - & - & \underline{0.901} & \underline{0.888} & \underline{0.906} \\
    \midrule
    CTransPath & SwinT* & $\sim$ 32K & 0.958 & 0.931 & 0.955 & 0.879 & 0.852 & 0.883 \\
    UNI & ViT-Large & $\sim$ 100K & - & - & - & - & 0.874 & 0.875    \\
    Virchow & ViT-Huge & $\sim$ 1.5M & \textbf{0.968} & \textbf{0.956} & \textbf{0.968} & \textbf{0.948} & \textbf{0.938} & \textbf{0.950}    \\
    \bottomrule
    \end{tabular}
    \label{tab:ab4}
\end{table*}

\subsection{Efficacy of the proposed methods} 
\label{sec:dis-pp}
In Section \ref{sec:pp}, we claim that the essence of the cross-scale positioning task is to bridge representations in a local view and those in a global view. To validate whether cross-scale positioning enhances models’ global understanding, we choose the WSI classification task since it is more related to a global understanding of a slide. We compare the performance of a purely MoCo v3 pretrained model using our dataset and the model using cross-scale positioning, named \emph{MoCo v3} and \emph{+XSP}, respectively. From the results in Table \ref{tab:ab1}, we can see that the latter model outperforms the purely MoCo v3 one in most metrics. The results can prove that cross-scale positioning boosts models' understanding of an image from a broader view. Besides, we also conduct ablating experiments on NCT-CRC-HE dataset, and observe a little performance gain after applying cross-scale positioning task as shown in Table \ref{tab:ab2}. The margin in full finetuning setting barely exists, which is reasonable since the two models share the same model architecture. If we take the results in Table \ref{tab:ab1} and Table \ref{tab:ab2} as a whole, we can derive the conclusion that cross-scale positioning helps model understand H\&E images better. 

In Section \ref{sec:exp-pdl1} and \ref{sec:exp-ihc}, we prove that our IHC model can surpass other pathological models, and we append an ablation study to further prove the efficacy of cross-stain transferring. We use the PD-L1 expression level assessing task. From the results in Table \ref{tab:ab3}, it is obvious that explicit transferring benefits the understanding of IHC images on the condition of a strong H\&E base, especially when the training date is limited.

\subsection{Data requirements of downstream tasks} 
\label{sec:dis-fdt} 
To prove the necessity of building up pathological foundation models, we further study the amounts of training data to effectively apply a foundation model to specific downstream tasks. 

\begin{figure}
	\centering
	\includegraphics[width=0.85\columnwidth]{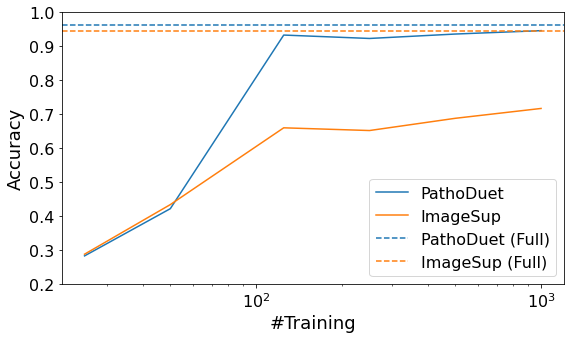}
	\caption{Data requirement study on NCT-CRC-HE dataset. PathoDuet is compared with ImageSup and the performance with the full dataset as an upper bound is represented by the dotted line with the same color.}
	\label{fig:req1}
\end{figure}

In Fig. \ref{fig:req1}, we compare our H\&E model with ImageNet-supervised ViT using the NCT-CRC-HE dataset. We can observe that although these two models can present close performance with the full dataset ($10^6$ images), when the amount is limited to only $10^2$, the performance gap is large. This can be used to validate that using pathological foundation models can reduce data requirement of downstream tasks in pathology.

\begin{figure}
	\centering
	\includegraphics[width=0.85\columnwidth]{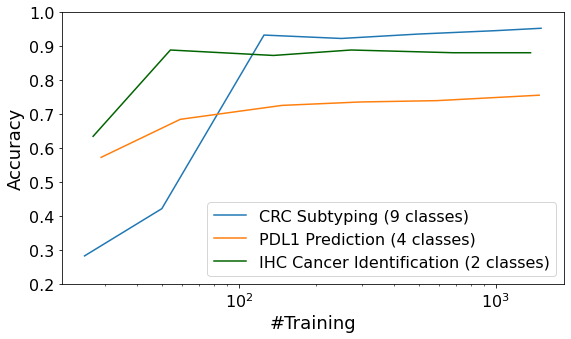}
	\caption{Data requirement study on different datasets using PathoDuet. }
	\label{fig:req2}
\end{figure}

To gain a more general insight in a direct way, we present the performance of PathoDuet over three different patch-level datasets, as introduced in Section \ref{sec:exp-crc}, \ref{sec:exp-pdl1} and \ref{sec:exp-ihc} respectively. From the results in Fig. \ref{fig:req2}, it is obvious that PathoDuet reduces the data requirements to about $10^2$ over different datasets. However, when there are only less than 50 training images, the models fail to tackle the downstream tasks and present a performance like random guess. 
Through these two experiments, we look further into the value of pathological foundation models in that they highly relieve the burden of data annotation.

\subsection{Comparison with giant pathological models} 
\label{sec:dis-paige} 
Recently, there has been a trend in pretraining pathological models with astronomical numbers of diagnostic slides. UNI, proposed by \citet{chen2023generalpurpose}, utilizes 100,000 slides to train a ViT-Large with DINO v2 framework. Virchow \citep{vorontsov2023virchow} further extends this number to 1.5 million and employs a ViT-Huge model with DINO v2 as well. In this section, we compare our model to those giants and CTransPath as a baseline in the CRC patch classification task. Notably, except our models, all the figures are copied from Virchow's experiment, so the results of CTransPath are different from that in Section \ref{sec:exp-crc}, which is as small as 0.002 in accuracy, and 0.001 in balanced accuracy and weight F1 score, and can be owed to different training strategies and randomness. Here, we provide two versions of our model, a normal ViT used throughout the aforementioned experiments, and a ViT with our positioner network to aggregate the features instead of using the vanilla average. The latter one is marked with a *.

Reviewing the results in Table \ref{tab:ab4}, we can see a small gap when using the normalized version of the dataset. The value of data is much more obvious with the non-normalized dataset since Virchow shows an impressive and dominant performance. However, the normal use of our 11K-slide-trained ViT-Base model can surpass the 100K-slide-trained UNI, which demonstrates the power of combining field knowledge. Hence, we claim that it is also of vital importance to employ a proper training strategy carefully tailored to pathological characteristics.

\section{Conclusion}
We introduce PathoDuet, a series of foundation models on computational pathology, covering both H\&E and IHC images, and propose a new self-supervised learning framework with two pretext tasks in pathology. The key to this framework is the introduction of a pretext token and following task raisers. It consists of both a model pretraining task, cross-scale positioning, and a model adaptation task, cross-stain transferring. In cross-scale positioning, we bridge the local and global representations of H\&E patches to enhance pathological image understanding in various magnifications. In cross-stain transferring, we utilize adaptive instance normalized H\&E features to provide pseudo-IHC features injected with structural information. The original H\&E model is therefore transferred to an interpreter of IHC images. We evaluate the performance of our models over a wide variety of downstream tasks, and the experimental results show the efficacy of our models on most tasks. Besides, we also investigate the downstream data requirements and comparison with giant pathological models, to discover the power of data and delicately designed SSL methods tailored to pathological images. PathoDuet highlights the importance of training strategy, while the giants, UNI and Virchow, point out the advantage of preparing sufficient training data. Hence, we will take all efforts to collect more data to iterate and upgrade our models in the future.

\section*{Acknowledgements}
This work was supported by National Natural Science Foundation of China (No.62301311).

\bibliographystyle{model2-names.bst}\biboptions{authoryear}
\bibliography{refs}

\begin{thebibliography}{62}
\expandafter\ifx\csname natexlab\endcsname\relax\def\natexlab#1{#1}\fi
\providecommand{\url}[1]{\texttt{#1}}
\providecommand{\href}[2]{#2}
\providecommand{\path}[1]{#1}
\providecommand{\DOIprefix}{doi:}
\providecommand{\ArXivprefix}{arXiv:}
\providecommand{\URLprefix}{URL: }
\providecommand{\Pubmedprefix}{pmid:}
\providecommand{\doi}[1]{\href{http://dx.doi.org/#1}{\path{#1}}}
\providecommand{\Pubmed}[1]{\href{pmid:#1}{\path{#1}}}
\providecommand{\bibinfo}[2]{#2}
\ifx\xfnm\relax \def\xfnm[#1]{\unskip,\space#1}\fi
\bibitem[{Abbet et~al.(2022)Abbet, Studer, Fischer, Dawson, Zlobec, Bozorgtabar and Thiran}]{abbet2022self}
\bibinfo{author}{Abbet, C.}, \bibinfo{author}{Studer, L.}, \bibinfo{author}{Fischer, A.}, \bibinfo{author}{Dawson, H.}, \bibinfo{author}{Zlobec, I.}, \bibinfo{author}{Bozorgtabar, B.}, \bibinfo{author}{Thiran, J.P.}, \bibinfo{year}{2022}.
\newblock \bibinfo{title}{Self-rule to multi-adapt: Generalized multi-source feature learning using unsupervised domain adaptation for colorectal cancer tissue detection}.
\newblock \bibinfo{journal}{Medical image analysis} \bibinfo{volume}{79}, \bibinfo{pages}{102473}.
\bibitem[{Aryal and Yahyasoltani(2023)}]{aryal2023context}
\bibinfo{author}{Aryal, M.}, \bibinfo{author}{Yahyasoltani, N.}, \bibinfo{year}{2023}.
\newblock \bibinfo{title}{Context-aware self-supervised learning of whole slide images}.
\newblock \bibinfo{journal}{arXiv preprint arXiv:2306.04763} .
\bibitem[{Bejnordi et~al.(2017)Bejnordi, Veta, Van~Diest, Van~Ginneken, Karssemeijer, Litjens, Van Der~Laak, Hermsen, Manson, Balkenhol et~al.}]{bejnordi2017diagnostic}
\bibinfo{author}{Bejnordi, B.E.}, \bibinfo{author}{Veta, M.}, \bibinfo{author}{Van~Diest, P.J.}, \bibinfo{author}{Van~Ginneken, B.}, \bibinfo{author}{Karssemeijer, N.}, \bibinfo{author}{Litjens, G.}, \bibinfo{author}{Van Der~Laak, J.A.}, \bibinfo{author}{Hermsen, M.}, \bibinfo{author}{Manson, Q.F.}, \bibinfo{author}{Balkenhol, M.}, et~al., \bibinfo{year}{2017}.
\newblock \bibinfo{title}{Diagnostic assessment of deep learning algorithms for detection of lymph node metastases in women with breast cancer}.
\newblock \bibinfo{journal}{Jama} \bibinfo{volume}{318}, \bibinfo{pages}{2199--2210}.
\bibitem[{Campanella et~al.(2019)Campanella, Hanna, Geneslaw, Miraflor, Werneck Krauss~Silva, Busam, Brogi, Reuter, Klimstra and Fuchs}]{campanella2019clinical}
\bibinfo{author}{Campanella, G.}, \bibinfo{author}{Hanna, M.G.}, \bibinfo{author}{Geneslaw, L.}, \bibinfo{author}{Miraflor, A.}, \bibinfo{author}{Werneck Krauss~Silva, V.}, \bibinfo{author}{Busam, K.J.}, \bibinfo{author}{Brogi, E.}, \bibinfo{author}{Reuter, V.E.}, \bibinfo{author}{Klimstra, D.S.}, \bibinfo{author}{Fuchs, T.J.}, \bibinfo{year}{2019}.
\newblock \bibinfo{title}{Clinical-grade computational pathology using weakly supervised deep learning on whole slide images}.
\newblock \bibinfo{journal}{Nature medicine} \bibinfo{volume}{25}, \bibinfo{pages}{1301--1309}.
\bibitem[{Caron et~al.(2021)Caron, Touvron, Misra, Jégou, Mairal, Bojanowski and Joulin}]{caron2021emerging}
\bibinfo{author}{Caron, M.}, \bibinfo{author}{Touvron, H.}, \bibinfo{author}{Misra, I.}, \bibinfo{author}{Jégou, H.}, \bibinfo{author}{Mairal, J.}, \bibinfo{author}{Bojanowski, P.}, \bibinfo{author}{Joulin, A.}, \bibinfo{year}{2021}.
\newblock \bibinfo{title}{Emerging properties in self-supervised vision transformers}.
\newblock \href{http://arxiv.org/abs/2104.14294}{\tt arXiv:2104.14294}.
\bibitem[{Chen et~al.(2023)Chen, Ding, Lu, Williamson, Jaume, Chen, Zhang, Shao, Song, Shaban, Williams, Vaidya, Sahai, Oldenburg, Weishaupt, Wang, Williams, Le, Gerber and Mahmood}]{chen2023generalpurpose}
\bibinfo{author}{Chen, R.J.}, \bibinfo{author}{Ding, T.}, \bibinfo{author}{Lu, M.Y.}, \bibinfo{author}{Williamson, D.F.K.}, \bibinfo{author}{Jaume, G.}, \bibinfo{author}{Chen, B.}, \bibinfo{author}{Zhang, A.}, \bibinfo{author}{Shao, D.}, \bibinfo{author}{Song, A.H.}, \bibinfo{author}{Shaban, M.}, \bibinfo{author}{Williams, M.}, \bibinfo{author}{Vaidya, A.}, \bibinfo{author}{Sahai, S.}, \bibinfo{author}{Oldenburg, L.}, \bibinfo{author}{Weishaupt, L.L.}, \bibinfo{author}{Wang, J.J.}, \bibinfo{author}{Williams, W.}, \bibinfo{author}{Le, L.P.}, \bibinfo{author}{Gerber, G.}, \bibinfo{author}{Mahmood, F.}, \bibinfo{year}{2023}.
\newblock \bibinfo{title}{A general-purpose self-supervised model for computational pathology}.
\newblock \href{http://arxiv.org/abs/2308.15474}{\tt arXiv:2308.15474}.
\bibitem[{Chen et~al.(2020)Chen, Kornblith, Norouzi and Hinton}]{chen2020simple}
\bibinfo{author}{Chen, T.}, \bibinfo{author}{Kornblith, S.}, \bibinfo{author}{Norouzi, M.}, \bibinfo{author}{Hinton, G.}, \bibinfo{year}{2020}.
\newblock \bibinfo{title}{A simple framework for contrastive learning of visual representations}, in: \bibinfo{booktitle}{International conference on machine learning (ICML)}, \bibinfo{organization}{PMLR}. pp. \bibinfo{pages}{1597--1607}.
\bibitem[{Chen and He(2021)}]{chen2021exploring}
\bibinfo{author}{Chen, X.}, \bibinfo{author}{He, K.}, \bibinfo{year}{2021}.
\newblock \bibinfo{title}{Exploring simple siamese representation learning}, in: \bibinfo{booktitle}{Proceedings of the IEEE/CVF conference on computer vision and pattern recognition (CVPR)}, pp. \bibinfo{pages}{15750--15758}.
\bibitem[{Chen et~al.(2021)Chen, Xie and He}]{chen2021empirical}
\bibinfo{author}{Chen, X.}, \bibinfo{author}{Xie, S.}, \bibinfo{author}{He, K.}, \bibinfo{year}{2021}.
\newblock \bibinfo{title}{An empirical study of training self-supervised vision transformers}, in: \bibinfo{booktitle}{Proceedings of the IEEE/CVF conference on computer vision and pattern recognition (CVPR)}, pp. \bibinfo{pages}{9640--9649}.
\bibitem[{Ciga et~al.(2022)Ciga, Xu and Martel}]{ciga2022self}
\bibinfo{author}{Ciga, O.}, \bibinfo{author}{Xu, T.}, \bibinfo{author}{Martel, A.L.}, \bibinfo{year}{2022}.
\newblock \bibinfo{title}{Self supervised contrastive learning for digital histopathology}.
\newblock \bibinfo{journal}{Machine learning with applications} \bibinfo{volume}{7}, \bibinfo{pages}{100198}.
\bibitem[{Deng et~al.(2009)Deng, Dong, Socher, Li, Li and Fei-Fei}]{5206848}
\bibinfo{author}{Deng, J.}, \bibinfo{author}{Dong, W.}, \bibinfo{author}{Socher, R.}, \bibinfo{author}{Li, L.J.}, \bibinfo{author}{Li, K.}, \bibinfo{author}{Fei-Fei, L.}, \bibinfo{year}{2009}.
\newblock \bibinfo{title}{Imagenet: A large-scale hierarchical image database}, in: \bibinfo{booktitle}{Proceedings of the IEEE/CVF conference on computer vision and pattern recognition (CVPR)}, pp. \bibinfo{pages}{248--255}.
\bibitem[{Dippel et~al.(2024)Dippel, Feulner, Winterhoff, Milbich, Tietz, Schallenberg, Dernbach, Kunft, Heinke, Eich, Ribbat-Idel, Krupar, Anders, Prenißl, Jurmeister, Horst, Ruff, Müller, Klauschen and Alber}]{dippel2024rudolfvfoundationmodelpathologists}
\bibinfo{author}{Dippel, J.}, \bibinfo{author}{Feulner, B.}, \bibinfo{author}{Winterhoff, T.}, \bibinfo{author}{Milbich, T.}, \bibinfo{author}{Tietz, S.}, \bibinfo{author}{Schallenberg, S.}, \bibinfo{author}{Dernbach, G.}, \bibinfo{author}{Kunft, A.}, \bibinfo{author}{Heinke, S.}, \bibinfo{author}{Eich, M.L.}, \bibinfo{author}{Ribbat-Idel, J.}, \bibinfo{author}{Krupar, R.}, \bibinfo{author}{Anders, P.}, \bibinfo{author}{Prenißl, N.}, \bibinfo{author}{Jurmeister, P.}, \bibinfo{author}{Horst, D.}, \bibinfo{author}{Ruff, L.}, \bibinfo{author}{Müller, K.R.}, \bibinfo{author}{Klauschen, F.}, \bibinfo{author}{Alber, M.}, \bibinfo{year}{2024}.
\newblock \bibinfo{title}{Rudolfv: A foundation model by pathologists for pathologists}.
\newblock \href{http://arxiv.org/abs/2401.04079}{\tt arXiv:2401.04079}.
\bibitem[{Doersch et~al.(2015)Doersch, Gupta and Efros}]{doersch2015unsupervised}
\bibinfo{author}{Doersch, C.}, \bibinfo{author}{Gupta, A.}, \bibinfo{author}{Efros, A.A.}, \bibinfo{year}{2015}.
\newblock \bibinfo{title}{Unsupervised visual representation learning by context prediction}, in: \bibinfo{booktitle}{Proceedings of the IEEE/CVF conference on computer vision and pattern recognition (CVPR)}, pp. \bibinfo{pages}{1422--1430}.
\bibitem[{Dosovitskiy et~al.(2020)Dosovitskiy, Beyer, Kolesnikov, Weissenborn, Zhai, Unterthiner, Dehghani, Minderer, Heigold, Gelly et~al.}]{dosovitskiy2020image}
\bibinfo{author}{Dosovitskiy, A.}, \bibinfo{author}{Beyer, L.}, \bibinfo{author}{Kolesnikov, A.}, \bibinfo{author}{Weissenborn, D.}, \bibinfo{author}{Zhai, X.}, \bibinfo{author}{Unterthiner, T.}, \bibinfo{author}{Dehghani, M.}, \bibinfo{author}{Minderer, M.}, \bibinfo{author}{Heigold, G.}, \bibinfo{author}{Gelly, S.}, et~al., \bibinfo{year}{2020}.
\newblock \bibinfo{title}{An image is worth 16x16 words: Transformers for image recognition at scale}.
\newblock \bibinfo{journal}{arXiv preprint arXiv:2010.11929} .
\bibitem[{Gidaris et~al.(2018)Gidaris, Singh and Komodakis}]{gidaris2018unsupervised}
\bibinfo{author}{Gidaris, S.}, \bibinfo{author}{Singh, P.}, \bibinfo{author}{Komodakis, N.}, \bibinfo{year}{2018}.
\newblock \bibinfo{title}{Unsupervised representation learning by predicting image rotations}.
\newblock \bibinfo{journal}{arXiv preprint arXiv:1803.07728} .
\bibitem[{Grill et~al.(2020)Grill, Strub, Altch{\'e}, Tallec, Richemond, Buchatskaya, Doersch, Avila~Pires, Guo, Gheshlaghi~Azar et~al.}]{grill2020bootstrap}
\bibinfo{author}{Grill, J.B.}, \bibinfo{author}{Strub, F.}, \bibinfo{author}{Altch{\'e}, F.}, \bibinfo{author}{Tallec, C.}, \bibinfo{author}{Richemond, P.}, \bibinfo{author}{Buchatskaya, E.}, \bibinfo{author}{Doersch, C.}, \bibinfo{author}{Avila~Pires, B.}, \bibinfo{author}{Guo, Z.}, \bibinfo{author}{Gheshlaghi~Azar, M.}, et~al., \bibinfo{year}{2020}.
\newblock \bibinfo{title}{Bootstrap your own latent-a new approach to self-supervised learning}.
\newblock \bibinfo{journal}{Advances in neural information processing systems} \bibinfo{volume}{33}, \bibinfo{pages}{21271--21284}.
\bibitem[{He et~al.(2022)He, Chen, Xie, Li, Doll{\'a}r and Girshick}]{he2022masked}
\bibinfo{author}{He, K.}, \bibinfo{author}{Chen, X.}, \bibinfo{author}{Xie, S.}, \bibinfo{author}{Li, Y.}, \bibinfo{author}{Doll{\'a}r, P.}, \bibinfo{author}{Girshick, R.}, \bibinfo{year}{2022}.
\newblock \bibinfo{title}{Masked autoencoders are scalable vision learners}, in: \bibinfo{booktitle}{Proceedings of the IEEE/CVF conference on computer vision and pattern recognition (CVPR)}, pp. \bibinfo{pages}{16000--16009}.
\bibitem[{He et~al.(2020)He, Fan, Wu, Xie and Girshick}]{he2020momentum}
\bibinfo{author}{He, K.}, \bibinfo{author}{Fan, H.}, \bibinfo{author}{Wu, Y.}, \bibinfo{author}{Xie, S.}, \bibinfo{author}{Girshick, R.}, \bibinfo{year}{2020}.
\newblock \bibinfo{title}{Momentum contrast for unsupervised visual representation learning}, in: \bibinfo{booktitle}{Proceedings of the IEEE/CVF conference on computer vision and pattern recognition (CVPR)}, pp. \bibinfo{pages}{9729--9738}.
\bibitem[{Huang and Belongie(2017)}]{huang2017arbitrary}
\bibinfo{author}{Huang, X.}, \bibinfo{author}{Belongie, S.}, \bibinfo{year}{2017}.
\newblock \bibinfo{title}{Arbitrary style transfer in real-time with adaptive instance normalization}, in: \bibinfo{booktitle}{Proceedings of the IEEE international conference on computer vision (ICCV)}, pp. \bibinfo{pages}{1501--1510}.
\bibitem[{Huang et~al.(2023a)Huang, Bianchi, Yuksekgonul, Montine and Zou}]{Huang2023PLIP}
\bibinfo{author}{Huang, Z.}, \bibinfo{author}{Bianchi, F.}, \bibinfo{author}{Yuksekgonul, M.}, \bibinfo{author}{Montine, T.}, \bibinfo{author}{Zou, J.}, \bibinfo{year}{2023}a.
\newblock \bibinfo{title}{Leveraging medical twitter to build a visual{\textendash}language foundation model for pathology ai}.
\newblock \bibinfo{journal}{bioRxiv} .
\bibitem[{Huang et~al.(2021)Huang, Chai, Wang, Wang, Yang and Wu}]{huang2021integration}
\bibinfo{author}{Huang, Z.}, \bibinfo{author}{Chai, H.}, \bibinfo{author}{Wang, R.}, \bibinfo{author}{Wang, H.}, \bibinfo{author}{Yang, Y.}, \bibinfo{author}{Wu, H.}, \bibinfo{year}{2021}.
\newblock \bibinfo{title}{Integration of patch features through self-supervised learning and transformer for survival analysis on whole slide images}, in: \bibinfo{booktitle}{Medical Image Computing and Computer Assisted Intervention (MICCAI)}, \bibinfo{organization}{Springer}. pp. \bibinfo{pages}{561--570}.
\bibitem[{Huang et~al.(2023b)Huang, Wang, Deng, Ye, Su, Sun, He, Gu, Gu, Zhang and Qiao}]{huang2023stunetscalabletransferablemedical}
\bibinfo{author}{Huang, Z.}, \bibinfo{author}{Wang, H.}, \bibinfo{author}{Deng, Z.}, \bibinfo{author}{Ye, J.}, \bibinfo{author}{Su, Y.}, \bibinfo{author}{Sun, H.}, \bibinfo{author}{He, J.}, \bibinfo{author}{Gu, Y.}, \bibinfo{author}{Gu, L.}, \bibinfo{author}{Zhang, S.}, \bibinfo{author}{Qiao, Y.}, \bibinfo{year}{2023}b.
\newblock \bibinfo{title}{Stu-net: Scalable and transferable medical image segmentation models empowered by large-scale supervised pre-training}.
\newblock \href{http://arxiv.org/abs/2304.06716}{\tt arXiv:2304.06716}.
\bibitem[{Jing and Tian(2020)}]{jing2020self}
\bibinfo{author}{Jing, L.}, \bibinfo{author}{Tian, Y.}, \bibinfo{year}{2020}.
\newblock \bibinfo{title}{Self-supervised visual feature learning with deep neural networks: A survey}.
\newblock \bibinfo{journal}{IEEE transactions on pattern analysis and machine intelligence (TPAMI)} \bibinfo{volume}{43}, \bibinfo{pages}{4037--4058}.
\bibitem[{Kather et~al.(2018)Kather, Halama and Marx}]{kather_jakob_nikolas_2018_1214456}
\bibinfo{author}{Kather, J.N.}, \bibinfo{author}{Halama, N.}, \bibinfo{author}{Marx, A.}, \bibinfo{year}{2018}.
\newblock \bibinfo{title}{{100,000 histological images of human colorectal cancer and healthy tissue}}.
\bibitem[{Kawai et~al.(2023)Kawai, Ota and Yamaoka}]{kawai2023large}
\bibinfo{author}{Kawai, M.}, \bibinfo{author}{Ota, N.}, \bibinfo{author}{Yamaoka, S.}, \bibinfo{year}{2023}.
\newblock \bibinfo{title}{Large-scale pretraining on pathological images for fine-tuning of small pathological benchmarks}.
\newblock \bibinfo{journal}{arXiv preprint arXiv:2303.15693} .
\bibitem[{Koohbanani et~al.(2021)Koohbanani, Unnikrishnan, Khurram, Krishnaswamy and Rajpoot}]{koohbanani2021self}
\bibinfo{author}{Koohbanani, N.A.}, \bibinfo{author}{Unnikrishnan, B.}, \bibinfo{author}{Khurram, S.A.}, \bibinfo{author}{Krishnaswamy, P.}, \bibinfo{author}{Rajpoot, N.}, \bibinfo{year}{2021}.
\newblock \bibinfo{title}{Self-path: Self-supervision for classification of pathology images with limited annotations}.
\newblock \bibinfo{journal}{IEEE Transactions on medical imaging} \bibinfo{volume}{40}, \bibinfo{pages}{2845--2856}.
\bibitem[{van~der Laak et~al.(2021)van~der Laak, Lotz, Weiss and Heldmann}]{pzj5-bs61-21}
\bibinfo{author}{van~der Laak, J.}, \bibinfo{author}{Lotz, J.}, \bibinfo{author}{Weiss, N.}, \bibinfo{author}{Heldmann, S.}, \bibinfo{year}{2021}.
\newblock \bibinfo{title}{Hyreco-hybrid re-stained and consecutive histological serial sections}.
\bibitem[{Lazard et~al.(2023)Lazard, Lerousseau, Decenci{\`e}re and Walter}]{lazard2023giga}
\bibinfo{author}{Lazard, T.}, \bibinfo{author}{Lerousseau, M.}, \bibinfo{author}{Decenci{\`e}re, E.}, \bibinfo{author}{Walter, T.}, \bibinfo{year}{2023}.
\newblock \bibinfo{title}{Giga-ssl: Self-supervised learning for gigapixel images}, in: \bibinfo{booktitle}{Proceedings of the IEEE/CVF Conference on Computer Vision and Pattern Recognition (CVPR)}, pp. \bibinfo{pages}{4304--4313}.
\bibitem[{Ledig et~al.(2017)Ledig, Theis, Husz{\'a}r, Caballero, Cunningham, Acosta, Aitken, Tejani, Totz, Wang et~al.}]{ledig2017photo}
\bibinfo{author}{Ledig, C.}, \bibinfo{author}{Theis, L.}, \bibinfo{author}{Husz{\'a}r, F.}, \bibinfo{author}{Caballero, J.}, \bibinfo{author}{Cunningham, A.}, \bibinfo{author}{Acosta, A.}, \bibinfo{author}{Aitken, A.}, \bibinfo{author}{Tejani, A.}, \bibinfo{author}{Totz, J.}, \bibinfo{author}{Wang, Z.}, et~al., \bibinfo{year}{2017}.
\newblock \bibinfo{title}{Photo-realistic single image super-resolution using a generative adversarial network}, in: \bibinfo{booktitle}{Proceedings of the IEEE/CVF conference on computer vision and pattern recognition (CVPR)}, pp. \bibinfo{pages}{4681--4690}.
\bibitem[{Li et~al.(2021)Li, Li and Eliceiri}]{li2021dual}
\bibinfo{author}{Li, B.}, \bibinfo{author}{Li, Y.}, \bibinfo{author}{Eliceiri, K.W.}, \bibinfo{year}{2021}.
\newblock \bibinfo{title}{Dual-stream multiple instance learning network for whole slide image classification with self-supervised contrastive learning}, in: \bibinfo{booktitle}{Proceedings of the IEEE/CVF conference on computer vision and pattern recognition (CVPR)}, pp. \bibinfo{pages}{14318--14328}.
\bibitem[{Li et~al.(2024)Li, Chen, Chu, Sun, Guan, Han and He}]{Li_2024_CVPR}
\bibinfo{author}{Li, J.}, \bibinfo{author}{Chen, Y.}, \bibinfo{author}{Chu, H.}, \bibinfo{author}{Sun, Q.}, \bibinfo{author}{Guan, T.}, \bibinfo{author}{Han, A.}, \bibinfo{author}{He, Y.}, \bibinfo{year}{2024}.
\newblock \bibinfo{title}{Dynamic graph representation with knowledge-aware attention for histopathology whole slide image analysis}, in: \bibinfo{booktitle}{Proceedings of the IEEE/CVF Conference on Computer Vision and Pattern Recognition (CVPR)}, pp. \bibinfo{pages}{11323--11332}.
\bibitem[{Li et~al.(2023)Li, Shang, Liu, Zhen, Zhu, Zhong, Sturgess, Zhou, Hu, Zhao, Wu, Li, Lin and Ren}]{Li2023}
\bibinfo{author}{Li, Z.}, \bibinfo{author}{Shang, Z.}, \bibinfo{author}{Liu, J.}, \bibinfo{author}{Zhen, H.}, \bibinfo{author}{Zhu, E.}, \bibinfo{author}{Zhong, S.}, \bibinfo{author}{Sturgess, R.N.}, \bibinfo{author}{Zhou, Y.}, \bibinfo{author}{Hu, X.}, \bibinfo{author}{Zhao, X.}, \bibinfo{author}{Wu, Y.}, \bibinfo{author}{Li, P.}, \bibinfo{author}{Lin, R.}, \bibinfo{author}{Ren, J.}, \bibinfo{year}{2023}.
\newblock \bibinfo{title}{D-lmbmap: a fully automated deep-learning pipeline for whole-brain profiling of neural circuitry}.
\newblock \bibinfo{journal}{Nature Methods} \bibinfo{volume}{20}, \bibinfo{pages}{1593–1604}.
\bibitem[{Ling et~al.(2023)Ling, Tan and Yan}]{ling2023self}
\bibinfo{author}{Ling, Y.}, \bibinfo{author}{Tan, W.}, \bibinfo{author}{Yan, B.}, \bibinfo{year}{2023}.
\newblock \bibinfo{title}{Self-supervised digital histopathology image disentanglement for arbitrary domain stain transfer}.
\newblock \bibinfo{journal}{IEEE Transactions on medical imaging} .
\bibitem[{Liu et~al.(2022)Liu, Zhu, Xu, Jia, Shi and Jin}]{Liu_2022_CVPR}
\bibinfo{author}{Liu, S.}, \bibinfo{author}{Zhu, C.}, \bibinfo{author}{Xu, F.}, \bibinfo{author}{Jia, X.}, \bibinfo{author}{Shi, Z.}, \bibinfo{author}{Jin, M.}, \bibinfo{year}{2022}.
\newblock \bibinfo{title}{Bci: Breast cancer immunohistochemical image generation through pyramid pix2pix}, in: \bibinfo{booktitle}{Proceedings of the IEEE/CVF Conference on Computer Vision and Pattern Recognition (CVPR) Workshops}, pp. \bibinfo{pages}{1815--1824}.
\bibitem[{Lotz et~al.(2022)Lotz, Weiss, van~der Laak and Heldmann}]{lotz2022comparison}
\bibinfo{author}{Lotz, J.}, \bibinfo{author}{Weiss, N.}, \bibinfo{author}{van~der Laak, J.}, \bibinfo{author}{Heldmann, S.}, \bibinfo{year}{2022}.
\newblock \bibinfo{title}{Comparison of consecutive and re-stained sections for image registration in histopathology}.
\newblock \href{http://arxiv.org/abs/2106.13150}{\tt arXiv:2106.13150}.
\bibitem[{Lu et~al.(2023)Lu, Chen, Williamson, Chen, Liang, Ding, Jaume, Odintsov, Zhang, Le et~al.}]{lu2023towards}
\bibinfo{author}{Lu, M.Y.}, \bibinfo{author}{Chen, B.}, \bibinfo{author}{Williamson, D.F.}, \bibinfo{author}{Chen, R.J.}, \bibinfo{author}{Liang, I.}, \bibinfo{author}{Ding, T.}, \bibinfo{author}{Jaume, G.}, \bibinfo{author}{Odintsov, I.}, \bibinfo{author}{Zhang, A.}, \bibinfo{author}{Le, L.P.}, et~al., \bibinfo{year}{2023}.
\newblock \bibinfo{title}{Towards a visual-language foundation model for computational pathology}.
\newblock \bibinfo{journal}{arXiv preprint arXiv:2307.12914} .
\bibitem[{Lu et~al.(2021)Lu, Williamson, Chen, Chen, Barbieri and Mahmood}]{lu2021data}
\bibinfo{author}{Lu, M.Y.}, \bibinfo{author}{Williamson, D.F.}, \bibinfo{author}{Chen, T.Y.}, \bibinfo{author}{Chen, R.J.}, \bibinfo{author}{Barbieri, M.}, \bibinfo{author}{Mahmood, F.}, \bibinfo{year}{2021}.
\newblock \bibinfo{title}{Data-efficient and weakly supervised computational pathology on whole-slide images}.
\newblock \bibinfo{journal}{Nature biomedical engineering} \bibinfo{volume}{5}, \bibinfo{pages}{555--570}.
\bibitem[{Ma et~al.(2024)Ma, Guo, Zhou, Wang, Xu, Cai, Zhu, Jin, Jiang, Han, Liang, Chan, Wang, Cheng and Chen}]{ma2024generalizablepathologyfoundationmodel}
\bibinfo{author}{Ma, J.}, \bibinfo{author}{Guo, Z.}, \bibinfo{author}{Zhou, F.}, \bibinfo{author}{Wang, Y.}, \bibinfo{author}{Xu, Y.}, \bibinfo{author}{Cai, Y.}, \bibinfo{author}{Zhu, Z.}, \bibinfo{author}{Jin, C.}, \bibinfo{author}{Jiang, Y.L.X.}, \bibinfo{author}{Han, A.}, \bibinfo{author}{Liang, L.}, \bibinfo{author}{Chan, R.C.K.}, \bibinfo{author}{Wang, J.}, \bibinfo{author}{Cheng, K.T.}, \bibinfo{author}{Chen, H.}, \bibinfo{year}{2024}.
\newblock \bibinfo{title}{Towards a generalizable pathology foundation model via unified knowledge distillation}.
\newblock \URLprefix \url{https://arxiv.org/abs/2407.18449}, \href{http://arxiv.org/abs/2407.18449}{\tt arXiv:2407.18449}.
\bibitem[{Noroozi and Favaro(2016)}]{noroozi2016unsupervised}
\bibinfo{author}{Noroozi, M.}, \bibinfo{author}{Favaro, P.}, \bibinfo{year}{2016}.
\newblock \bibinfo{title}{Unsupervised learning of visual representations by solving jigsaw puzzles}, in: \bibinfo{booktitle}{The European Conference on Computer Vision (ECCV)}, \bibinfo{organization}{Springer}. pp. \bibinfo{pages}{69--84}.
\bibitem[{Oquab et~al.(2023)Oquab, Darcet, Moutakanni, Vo, Szafraniec, Khalidov, Fernandez, Haziza, Massa, El-Nouby, Assran, Ballas, Galuba, Howes, Huang, Li, Misra, Rabbat, Sharma, Synnaeve, Xu, Jegou, Mairal, Labatut, Joulin and Bojanowski}]{oquab2023dinov2}
\bibinfo{author}{Oquab, M.}, \bibinfo{author}{Darcet, T.}, \bibinfo{author}{Moutakanni, T.}, \bibinfo{author}{Vo, H.}, \bibinfo{author}{Szafraniec, M.}, \bibinfo{author}{Khalidov, V.}, \bibinfo{author}{Fernandez, P.}, \bibinfo{author}{Haziza, D.}, \bibinfo{author}{Massa, F.}, \bibinfo{author}{El-Nouby, A.}, \bibinfo{author}{Assran, M.}, \bibinfo{author}{Ballas, N.}, \bibinfo{author}{Galuba, W.}, \bibinfo{author}{Howes, R.}, \bibinfo{author}{Huang, P.Y.}, \bibinfo{author}{Li, S.W.}, \bibinfo{author}{Misra, I.}, \bibinfo{author}{Rabbat, M.}, \bibinfo{author}{Sharma, V.}, \bibinfo{author}{Synnaeve, G.}, \bibinfo{author}{Xu, H.}, \bibinfo{author}{Jegou, H.}, \bibinfo{author}{Mairal, J.}, \bibinfo{author}{Labatut, P.}, \bibinfo{author}{Joulin, A.}, \bibinfo{author}{Bojanowski, P.}, \bibinfo{year}{2023}.
\newblock \bibinfo{title}{Dinov2: Learning robust visual features without supervision}.
\newblock \href{http://arxiv.org/abs/2304.07193}{\tt arXiv:2304.07193}.
\bibitem[{Pathak et~al.(2016)Pathak, Krahenbuhl, Donahue, Darrell and Efros}]{pathak2016context}
\bibinfo{author}{Pathak, D.}, \bibinfo{author}{Krahenbuhl, P.}, \bibinfo{author}{Donahue, J.}, \bibinfo{author}{Darrell, T.}, \bibinfo{author}{Efros, A.A.}, \bibinfo{year}{2016}.
\newblock \bibinfo{title}{Context encoders: Feature learning by inpainting}, in: \bibinfo{booktitle}{Proceedings of the IEEE/CVF conference on computer vision and pattern recognition (CVPR)}, pp. \bibinfo{pages}{2536--2544}.
\bibitem[{Pisula and Bozek(2022)}]{pisula2022language}
\bibinfo{author}{Pisula, J.I.}, \bibinfo{author}{Bozek, K.}, \bibinfo{year}{2022}.
\newblock \bibinfo{title}{Language models are good pathologists: using attention-based sequence reduction and text-pretrained transformers for efficient wsi classification}.
\newblock \bibinfo{journal}{arXiv preprint arXiv:2211.07384} .
\bibitem[{Sahasrabudhe et~al.(2020)Sahasrabudhe, Christodoulidis, Salgado, Michiels, Loi, Andr{\'e}, Paragios and Vakalopoulou}]{sahasrabudhe2020self}
\bibinfo{author}{Sahasrabudhe, M.}, \bibinfo{author}{Christodoulidis, S.}, \bibinfo{author}{Salgado, R.}, \bibinfo{author}{Michiels, S.}, \bibinfo{author}{Loi, S.}, \bibinfo{author}{Andr{\'e}, F.}, \bibinfo{author}{Paragios, N.}, \bibinfo{author}{Vakalopoulou, M.}, \bibinfo{year}{2020}.
\newblock \bibinfo{title}{Self-supervised nuclei segmentation in histopathological images using attention}, in: \bibinfo{booktitle}{Medical Image Computing and Computer Assisted Intervention (MICCAI)}, \bibinfo{organization}{Springer}. pp. \bibinfo{pages}{393--402}.
\bibitem[{Schirris et~al.(2022)Schirris, Gavves, Nederlof, Horlings and Teuwen}]{schirris2022deepsmile}
\bibinfo{author}{Schirris, Y.}, \bibinfo{author}{Gavves, E.}, \bibinfo{author}{Nederlof, I.}, \bibinfo{author}{Horlings, H.M.}, \bibinfo{author}{Teuwen, J.}, \bibinfo{year}{2022}.
\newblock \bibinfo{title}{Deepsmile: contrastive self-supervised pre-training benefits msi and hrd classification directly from h\&e whole-slide images in colorectal and breast cancer}.
\newblock \bibinfo{journal}{Medical image analysis} \bibinfo{volume}{79}, \bibinfo{pages}{102464}.
\bibitem[{Shao et~al.(2021)Shao, Bian, Chen, Wang, Zhang, Ji et~al.}]{shao2021transmil}
\bibinfo{author}{Shao, Z.}, \bibinfo{author}{Bian, H.}, \bibinfo{author}{Chen, Y.}, \bibinfo{author}{Wang, Y.}, \bibinfo{author}{Zhang, J.}, \bibinfo{author}{Ji, X.}, et~al., \bibinfo{year}{2021}.
\newblock \bibinfo{title}{Transmil: Transformer based correlated multiple instance learning for whole slide image classification}.
\newblock \bibinfo{journal}{Advances in neural information processing systems} \bibinfo{volume}{34}, \bibinfo{pages}{2136--2147}.
\bibitem[{Shekhar et~al.(2023)Shekhar, Bordes, Vincent and Morcos}]{shekhar2023objectives}
\bibinfo{author}{Shekhar, S.}, \bibinfo{author}{Bordes, F.}, \bibinfo{author}{Vincent, P.}, \bibinfo{author}{Morcos, A.}, \bibinfo{year}{2023}.
\newblock \bibinfo{title}{Objectives matter: Understanding the impact of self-supervised objectives on vision transformer representations}.
\newblock \bibinfo{journal}{arXiv preprint arXiv:2304.13089} .
\bibitem[{Srinidhi et~al.(2022)Srinidhi, Kim, Chen and Martel}]{srinidhi2022self}
\bibinfo{author}{Srinidhi, C.L.}, \bibinfo{author}{Kim, S.W.}, \bibinfo{author}{Chen, F.D.}, \bibinfo{author}{Martel, A.L.}, \bibinfo{year}{2022}.
\newblock \bibinfo{title}{Self-supervised driven consistency training for annotation efficient histopathology image analysis}.
\newblock \bibinfo{journal}{Medical image analysis} \bibinfo{volume}{75}, \bibinfo{pages}{102256}.
\bibitem[{Tiu et~al.(2022)Tiu, Talius, Patel, Langlotz, Ng and Rajpurkar}]{tiu2022expert}
\bibinfo{author}{Tiu, E.}, \bibinfo{author}{Talius, E.}, \bibinfo{author}{Patel, P.}, \bibinfo{author}{Langlotz, C.P.}, \bibinfo{author}{Ng, A.Y.}, \bibinfo{author}{Rajpurkar, P.}, \bibinfo{year}{2022}.
\newblock \bibinfo{title}{Expert-level detection of pathologies from unannotated chest x-ray images via self-supervised learning}.
\newblock \bibinfo{journal}{Nature Biomedical Engineering} \bibinfo{volume}{6}, \bibinfo{pages}{1399--1406}.
\bibitem[{Vorontsov et~al.(2023)Vorontsov, Bozkurt, Casson, Shaikovski, Zelechowski, Liu, Mathieu, van Eck, Lee, Viret, Robert, Wang, Kunz, Lee, Bernhard, Godrich, Oakley, Millar, Hanna, Retamero, Moye, Yousfi, Kanan, Klimstra, Rothrock and Fuchs}]{vorontsov2023virchow}
\bibinfo{author}{Vorontsov, E.}, \bibinfo{author}{Bozkurt, A.}, \bibinfo{author}{Casson, A.}, \bibinfo{author}{Shaikovski, G.}, \bibinfo{author}{Zelechowski, M.}, \bibinfo{author}{Liu, S.}, \bibinfo{author}{Mathieu, P.}, \bibinfo{author}{van Eck, A.}, \bibinfo{author}{Lee, D.}, \bibinfo{author}{Viret, J.}, \bibinfo{author}{Robert, E.}, \bibinfo{author}{Wang, Y.K.}, \bibinfo{author}{Kunz, J.D.}, \bibinfo{author}{Lee, M.C.H.}, \bibinfo{author}{Bernhard, J.}, \bibinfo{author}{Godrich, R.A.}, \bibinfo{author}{Oakley, G.}, \bibinfo{author}{Millar, E.}, \bibinfo{author}{Hanna, M.}, \bibinfo{author}{Retamero, J.}, \bibinfo{author}{Moye, W.A.}, \bibinfo{author}{Yousfi, R.}, \bibinfo{author}{Kanan, C.}, \bibinfo{author}{Klimstra, D.}, \bibinfo{author}{Rothrock, B.}, \bibinfo{author}{Fuchs, T.J.}, \bibinfo{year}{2023}.
\newblock \bibinfo{title}{Virchow: A million-slide digital pathology foundation model}.
\newblock \href{http://arxiv.org/abs/2309.07778}{\tt arXiv:2309.07778}.
\bibitem[{Vu et~al.(2023)Vu, Rajpoot, Raza and Rajpoot}]{vu2023handcrafted}
\bibinfo{author}{Vu, Q.D.}, \bibinfo{author}{Rajpoot, K.}, \bibinfo{author}{Raza, S.E.A.}, \bibinfo{author}{Rajpoot, N.}, \bibinfo{year}{2023}.
\newblock \bibinfo{title}{Handcrafted histological transformer (h2t): Unsupervised representation of whole slide images}.
\newblock \bibinfo{journal}{Medical image analysis} \bibinfo{volume}{85}, \bibinfo{pages}{102743}.
\bibitem[{Wang et~al.(2023a)Wang, Ahn and Kim}]{wang2023dual}
\bibinfo{author}{Wang, H.}, \bibinfo{author}{Ahn, E.}, \bibinfo{author}{Kim, J.}, \bibinfo{year}{2023}a.
\newblock \bibinfo{title}{A dual-branch self-supervised representation learning framework for tumour segmentation in whole slide images}.
\newblock \bibinfo{journal}{arXiv preprint arXiv:2303.11019} .
\bibitem[{Wang et~al.(2023b)Wang, Du, Yang, Zhang, Wang, Zhang, Yang, Huang and Han}]{wang2023retccl}
\bibinfo{author}{Wang, X.}, \bibinfo{author}{Du, Y.}, \bibinfo{author}{Yang, S.}, \bibinfo{author}{Zhang, J.}, \bibinfo{author}{Wang, M.}, \bibinfo{author}{Zhang, J.}, \bibinfo{author}{Yang, W.}, \bibinfo{author}{Huang, J.}, \bibinfo{author}{Han, X.}, \bibinfo{year}{2023}b.
\newblock \bibinfo{title}{Retccl: clustering-guided contrastive learning for whole-slide image retrieval}.
\newblock \bibinfo{journal}{Medical image analysis} \bibinfo{volume}{83}, \bibinfo{pages}{102645}.
\bibitem[{Wang et~al.(2022)Wang, Yang, Zhang, Wang, Zhang, Yang, Huang and Han}]{wang2022transformer}
\bibinfo{author}{Wang, X.}, \bibinfo{author}{Yang, S.}, \bibinfo{author}{Zhang, J.}, \bibinfo{author}{Wang, M.}, \bibinfo{author}{Zhang, J.}, \bibinfo{author}{Yang, W.}, \bibinfo{author}{Huang, J.}, \bibinfo{author}{Han, X.}, \bibinfo{year}{2022}.
\newblock \bibinfo{title}{Transformer-based unsupervised contrastive learning for histopathological image classification}.
\newblock \bibinfo{journal}{Medical image analysis} \bibinfo{volume}{81}, \bibinfo{pages}{102559}.
\bibitem[{Wang et~al.(2024)Wang, Liu, Zhang and Dou}]{wang2024foundationmodelendoscopyvideo}
\bibinfo{author}{Wang, Z.}, \bibinfo{author}{Liu, C.}, \bibinfo{author}{Zhang, S.}, \bibinfo{author}{Dou, Q.}, \bibinfo{year}{2024}.
\newblock \bibinfo{title}{Foundation model for endoscopy video analysis via large-scale self-supervised pre-train}.
\newblock \URLprefix \url{https://arxiv.org/abs/2306.16741}, \href{http://arxiv.org/abs/2306.16741}{\tt arXiv:2306.16741}.
\bibitem[{Xu et~al.(2024a)Xu, Usuyama, Bagga, Zhang, Rao, Naumann, Wong, Gero, González, Gu, Xu, Wei, Wang, Ma, Wei, Yang, Li, Gao, Rosemon, Bower, Lee, Weerasinghe, Wright, Robicsek, Piening, Bifulco, Wang and Poon}]{Xu2024}
\bibinfo{author}{Xu, H.}, \bibinfo{author}{Usuyama, N.}, \bibinfo{author}{Bagga, J.}, \bibinfo{author}{Zhang, S.}, \bibinfo{author}{Rao, R.}, \bibinfo{author}{Naumann, T.}, \bibinfo{author}{Wong, C.}, \bibinfo{author}{Gero, Z.}, \bibinfo{author}{González, J.}, \bibinfo{author}{Gu, Y.}, \bibinfo{author}{Xu, Y.}, \bibinfo{author}{Wei, M.}, \bibinfo{author}{Wang, W.}, \bibinfo{author}{Ma, S.}, \bibinfo{author}{Wei, F.}, \bibinfo{author}{Yang, J.}, \bibinfo{author}{Li, C.}, \bibinfo{author}{Gao, J.}, \bibinfo{author}{Rosemon, J.}, \bibinfo{author}{Bower, T.}, \bibinfo{author}{Lee, S.}, \bibinfo{author}{Weerasinghe, R.}, \bibinfo{author}{Wright, B.J.}, \bibinfo{author}{Robicsek, A.}, \bibinfo{author}{Piening, B.}, \bibinfo{author}{Bifulco, C.}, \bibinfo{author}{Wang, S.}, \bibinfo{author}{Poon, H.}, \bibinfo{year}{2024}a.
\newblock \bibinfo{title}{A whole-slide foundation model for digital pathology from real-world data}.
\newblock \bibinfo{journal}{Nature} \bibinfo{volume}{630}, \bibinfo{pages}{181–188}.
\bibitem[{Xu et~al.(2024b)Xu, Wang, Zhou, Ma, Yang, Lin, Wang, Wang, Liang, Han, Chan and Chen}]{xu2024multimodalknowledgeenhancedwholeslidepathology}
\bibinfo{author}{Xu, Y.}, \bibinfo{author}{Wang, Y.}, \bibinfo{author}{Zhou, F.}, \bibinfo{author}{Ma, J.}, \bibinfo{author}{Yang, S.}, \bibinfo{author}{Lin, H.}, \bibinfo{author}{Wang, X.}, \bibinfo{author}{Wang, J.}, \bibinfo{author}{Liang, L.}, \bibinfo{author}{Han, A.}, \bibinfo{author}{Chan, R.C.K.}, \bibinfo{author}{Chen, H.}, \bibinfo{year}{2024}b.
\newblock \bibinfo{title}{A multimodal knowledge-enhanced whole-slide pathology foundation model}.
\newblock \href{http://arxiv.org/abs/2407.15362}{\tt arXiv:2407.15362}.
\bibitem[{Yang et~al.(2022)Yang, Yin, Lu, Hu, Zhang, Jiang and Lv}]{yang2022cs}
\bibinfo{author}{Yang, P.}, \bibinfo{author}{Yin, X.}, \bibinfo{author}{Lu, H.}, \bibinfo{author}{Hu, Z.}, \bibinfo{author}{Zhang, X.}, \bibinfo{author}{Jiang, R.}, \bibinfo{author}{Lv, H.}, \bibinfo{year}{2022}.
\newblock \bibinfo{title}{Cs-co: A hybrid self-supervised visual representation learning method for h\&e-stained histopathological images}.
\newblock \bibinfo{journal}{Medical image analysis} \bibinfo{volume}{81}, \bibinfo{pages}{102539}.
\bibitem[{Zhang et~al.(2016)Zhang, Isola and Efros}]{zhang2016colorful}
\bibinfo{author}{Zhang, R.}, \bibinfo{author}{Isola, P.}, \bibinfo{author}{Efros, A.A.}, \bibinfo{year}{2016}.
\newblock \bibinfo{title}{Colorful image colorization}, in: \bibinfo{booktitle}{The European Conference on Computer Vision (ECCV)}, \bibinfo{organization}{Springer}. pp. \bibinfo{pages}{649--666}.
\bibitem[{Zhang and Metaxas(2024)}]{ZHANG2024102996}
\bibinfo{author}{Zhang, S.}, \bibinfo{author}{Metaxas, D.}, \bibinfo{year}{2024}.
\newblock \bibinfo{title}{On the challenges and perspectives of foundation models for medical image analysis}.
\newblock \bibinfo{journal}{Medical Image Analysis} \bibinfo{volume}{91}, \bibinfo{pages}{102996}.
\bibitem[{Zhao et~al.(2022)Zhao, Han, Pan, Lin, Yi, Liang, Chen, Li, Qiu, Li et~al.}]{zhao2022restainnet}
\bibinfo{author}{Zhao, B.}, \bibinfo{author}{Han, C.}, \bibinfo{author}{Pan, X.}, \bibinfo{author}{Lin, J.}, \bibinfo{author}{Yi, Z.}, \bibinfo{author}{Liang, C.}, \bibinfo{author}{Chen, X.}, \bibinfo{author}{Li, B.}, \bibinfo{author}{Qiu, W.}, \bibinfo{author}{Li, D.}, et~al., \bibinfo{year}{2022}.
\newblock \bibinfo{title}{Restainnet: a self-supervised digital re-stainer for stain normalization}.
\newblock \bibinfo{journal}{Computers and electrical engineering} \bibinfo{volume}{103}, \bibinfo{pages}{108304}.
\bibitem[{Zhou et~al.(2023)Zhou, Chia, Wagner, Ayhan, Williamson, Struyven, Liu, Xu, Lozano, Woodward-Court, Kihara, Allen, Gallacher, Littlejohns, Aslam, Bishop, Black, Sergouniotis, Atan, Dick, Williams, Barman, Barrett, Mackie, Braithwaite, Carare, Ennis, Gibson, Lotery, Self, Chakravarthy, Hogg, Paterson, Woodside, Peto, Mckay, Mcguinness, Foster, Balaskas, Khawaja, Pontikos, Rahi, Lascaratos, Patel, Chan, Chua, Day, Desai, Egan, Fruttiger, Garway-Heath, Hardcastle, Khaw, Moore, Sivaprasad, Strouthidis, Thomas, Tufail, Viswanathan, Dhillon, Macgillivray, Sudlow, Vitart, Doney, Trucco, Guggeinheim, Morgan, Hammond, Williams, Hysi, Harding, Zheng, Luben, Luthert, Sun, McKibbin, O’Sullivan, Oram, Weedon, Owen, Rudnicka, Sattar, Steel, Stratton, Tapp, Yates, Petzold, Madhusudhan, Altmann, Lee, Topol, Denniston, Alexander and Keane}]{Zhou2023}
\bibinfo{author}{Zhou, Y.}, \bibinfo{author}{Chia, M.A.}, \bibinfo{author}{Wagner, S.K.}, \bibinfo{author}{Ayhan, M.S.}, \bibinfo{author}{Williamson, D.J.}, \bibinfo{author}{Struyven, R.R.}, \bibinfo{author}{Liu, T.}, \bibinfo{author}{Xu, M.}, \bibinfo{author}{Lozano, M.G.}, \bibinfo{author}{Woodward-Court, P.}, \bibinfo{author}{Kihara, Y.}, \bibinfo{author}{Allen, N.}, \bibinfo{author}{Gallacher, J.E.J.}, \bibinfo{author}{Littlejohns, T.}, \bibinfo{author}{Aslam, T.}, \bibinfo{author}{Bishop, P.}, \bibinfo{author}{Black, G.}, \bibinfo{author}{Sergouniotis, P.}, \bibinfo{author}{Atan, D.}, \bibinfo{author}{Dick, A.D.}, \bibinfo{author}{Williams, C.}, \bibinfo{author}{Barman, S.}, \bibinfo{author}{Barrett, J.H.}, \bibinfo{author}{Mackie, S.}, \bibinfo{author}{Braithwaite, T.}, \bibinfo{author}{Carare, R.O.}, \bibinfo{author}{Ennis, S.}, \bibinfo{author}{Gibson, J.}, \bibinfo{author}{Lotery, A.J.}, \bibinfo{author}{Self, J.}, \bibinfo{author}{Chakravarthy, U.}, \bibinfo{author}{Hogg, R.E.},
  \bibinfo{author}{Paterson, E.}, \bibinfo{author}{Woodside, J.}, \bibinfo{author}{Peto, T.}, \bibinfo{author}{Mckay, G.}, \bibinfo{author}{Mcguinness, B.}, \bibinfo{author}{Foster, P.J.}, \bibinfo{author}{Balaskas, K.}, \bibinfo{author}{Khawaja, A.P.}, \bibinfo{author}{Pontikos, N.}, \bibinfo{author}{Rahi, J.S.}, \bibinfo{author}{Lascaratos, G.}, \bibinfo{author}{Patel, P.J.}, \bibinfo{author}{Chan, M.}, \bibinfo{author}{Chua, S.Y.L.}, \bibinfo{author}{Day, A.}, \bibinfo{author}{Desai, P.}, \bibinfo{author}{Egan, C.}, \bibinfo{author}{Fruttiger, M.}, \bibinfo{author}{Garway-Heath, D.F.}, \bibinfo{author}{Hardcastle, A.}, \bibinfo{author}{Khaw, S.P.T.}, \bibinfo{author}{Moore, T.}, \bibinfo{author}{Sivaprasad, S.}, \bibinfo{author}{Strouthidis, N.}, \bibinfo{author}{Thomas, D.}, \bibinfo{author}{Tufail, A.}, \bibinfo{author}{Viswanathan, A.C.}, \bibinfo{author}{Dhillon, B.}, \bibinfo{author}{Macgillivray, T.}, \bibinfo{author}{Sudlow, C.}, \bibinfo{author}{Vitart, V.}, \bibinfo{author}{Doney, A.},
  \bibinfo{author}{Trucco, E.}, \bibinfo{author}{Guggeinheim, J.A.}, \bibinfo{author}{Morgan, J.E.}, \bibinfo{author}{Hammond, C.J.}, \bibinfo{author}{Williams, K.}, \bibinfo{author}{Hysi, P.}, \bibinfo{author}{Harding, S.P.}, \bibinfo{author}{Zheng, Y.}, \bibinfo{author}{Luben, R.}, \bibinfo{author}{Luthert, P.}, \bibinfo{author}{Sun, Z.}, \bibinfo{author}{McKibbin, M.}, \bibinfo{author}{O’Sullivan, E.}, \bibinfo{author}{Oram, R.}, \bibinfo{author}{Weedon, M.}, \bibinfo{author}{Owen, C.G.}, \bibinfo{author}{Rudnicka, A.R.}, \bibinfo{author}{Sattar, N.}, \bibinfo{author}{Steel, D.}, \bibinfo{author}{Stratton, I.}, \bibinfo{author}{Tapp, R.}, \bibinfo{author}{Yates, M.M.}, \bibinfo{author}{Petzold, A.}, \bibinfo{author}{Madhusudhan, S.}, \bibinfo{author}{Altmann, A.}, \bibinfo{author}{Lee, A.Y.}, \bibinfo{author}{Topol, E.J.}, \bibinfo{author}{Denniston, A.K.}, \bibinfo{author}{Alexander, D.C.}, \bibinfo{author}{Keane, P.A.}, \bibinfo{year}{2023}.
\newblock \bibinfo{title}{A foundation model for generalizable disease detection from retinal images}.
\newblock \bibinfo{journal}{Nature} \bibinfo{volume}{622}, \bibinfo{pages}{156–163}.
\bibitem[{Zhu et~al.(2017)Zhu, Park, Isola and Efros}]{zhu2017unpaired}
\bibinfo{author}{Zhu, J.Y.}, \bibinfo{author}{Park, T.}, \bibinfo{author}{Isola, P.}, \bibinfo{author}{Efros, A.A.}, \bibinfo{year}{2017}.
\newblock \bibinfo{title}{Unpaired image-to-image translation using cycle-consistent adversarial networks}, in: \bibinfo{booktitle}{Proceedings of the IEEE international conference on computer vision (ICCV)}, pp. \bibinfo{pages}{2223--2232}.

\end{thebibliography}

\end{document}